\documentclass[sigconf]{acmart}
\usepackage{booktabs} 
\usepackage{color}
\definecolor{ao}{rgb}{0.0, 0.5, 0.0}
\usepackage{comment}
\usepackage{booktabs}
\usepackage{amsfonts}
\usepackage{amsmath}

\usepackage{amssymb}
\usepackage{makecell}
\usepackage{subcaption}
\usepackage{multirow}
\usepackage{pbox}
\usepackage{hhline}
\usepackage{graphicx}
\usepackage[flushleft]{threeparttable}
\usepackage{svg}
\usepackage{courier}
\usepackage{xspace}
\usepackage{epstopdf}
\usepackage{stfloats}
\usepackage{array}
\usepackage{xcolor,colortbl}
\colorlet{shadecolor}{gray!20}

\usepackage[aboveskip=1pt]{subcaption}
\usepackage[skip=0.85\baselineskip]{caption}
\usepackage{makecell}
\newcommand\newtoprule{\Xhline{.08em}}
\newcommand\newmidrule{\Xhline{.05em}}
\newcommand\newbottomrule{\Xhline{.08em}}

\newcommand{\modelnamens}{JOIE}
\newcommand{\modelname}{{\modelnamens}}
\newcommand{\tabincell}[2]{\begin{tabular}{@{}#1@{}}#2\end{tabular}}

\newcommand{\nop}[1]{}

\copyrightyear{2019} 
\acmYear{2019} 
\setcopyright{acmlicensed}
\acmConference[KDD '19]{The 25th ACM SIGKDD Conference on Knowledge Discovery and Data Mining}{August 4--8, 2019}{Anchorage, AK, USA}
\acmBooktitle{The 25th ACM SIGKDD Conference on Knowledge Discovery and Data Mining (KDD '19), August 4--8, 2019, Anchorage, AK, USA}
\acmPrice{15.00}
\acmDOI{10.1145/3292500.3330838}
\acmISBN{978-1-4503-6201-6/19/08}

\begin{document}
\settopmatter{printacmref=true}
\fancyhead{}
\title[Research Track Paper]{Universal Representation Learning of Knowledge Bases by Jointly Embedding Instances and Ontological Concepts}


\author{Junheng Hao, Muhao Chen, Wenchao Yu, Yizhou Sun, Wei Wang}
\affiliation{%
  \institution{University of California, Los Angeles}
}
\email{{jhao,yzsun, weiwang}@cs.ucla.edu, {muhaochen, yuwenchao}@ucla.edu}

\renewcommand{\shortauthors}{Research Track Paper}

\begin{abstract}
Many large-scale knowledge bases simultaneously represent two views of knowledge graphs (KGs): an \emph{ontology view} for abstract and commonsense concepts, and an \emph{instance view} for specific entities that are instantiated from ontological concepts.
Existing KG embedding models, however, merely focus on representing one of the two views alone. 
In this paper, we propose a novel two-view KG embedding model, \texttt{\modelname}, with the goal to produce better knowledge embedding 
and enable new applications that rely on multi-view knowledge.
\texttt{\modelname} employs both cross-view and intra-view modeling that learn on multiple facets of the knowledge base.
The cross-view association model is learned to bridge the embeddings of ontological concepts and their corresponding instance-view entities. The intra-view models are trained to capture the structured knowledge of instance and ontology views in separate embedding spaces, with a hierarchy-aware encoding technique enabled for ontologies with hierarchies. 
We explore multiple representation techniques for the two model components and investigate with nine variants of \texttt{\modelname}.
Our model is trained on large-scale knowledge bases that consist of massive instances and their corresponding ontological concepts connected via a (small) set of cross-view links.
Experimental results on public datasets show that the best variant of \texttt{\modelname} significantly outperforms previous models on instance-view triple prediction task as well as ontology population on ontology-view KG.
In addition, our model successfully extends 
the use of KG embeddings to entity typing with promising performance.\par

\end{abstract}

%
%


\maketitle

\section{Introduction}\label{sec:introduction}
\begin{figure}[htbp]
	\centering
	\includegraphics[width=\columnwidth]{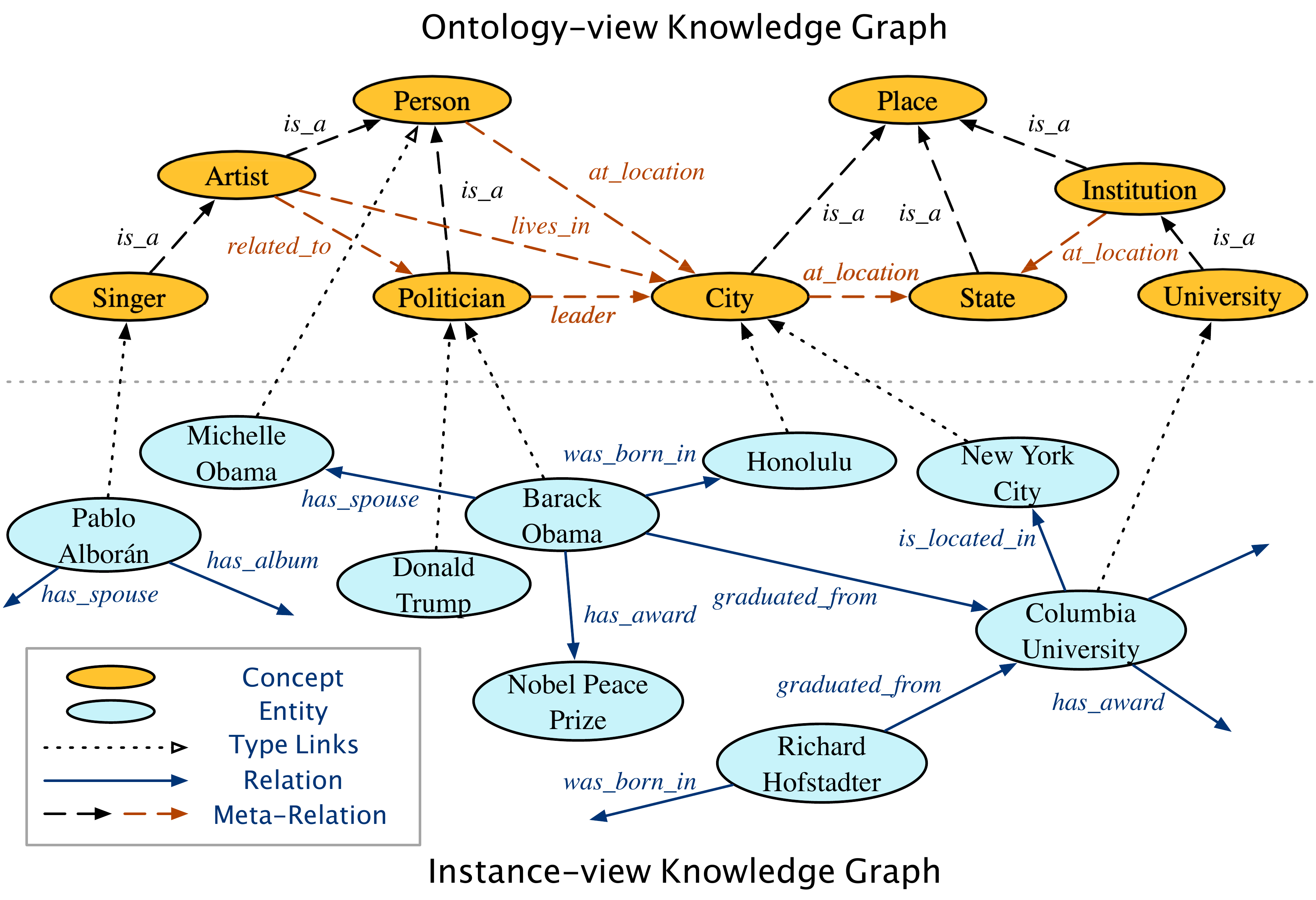}
	\caption{An example of two-view KB. Regular meta-relations and  hierarchical meta-relations are denoted as orange and black dashed lines respectively in the ontology view.
	}
	\label{fig:two-view}
\end{figure}

Knowledge bases (KBs), such as DBpedia~\cite{lehmann2015dbpedia}, YAGO~\cite{mahdisoltani2014yago3} and ConceptNet~\cite{speer2017conceptnet}, have incorporated large-scale multi-relational data and motivated many knowledge-driven applications.
These KBs store knowledge graphs (KGs) that can be categorized as two views:
(i) the \textbf{instance-view knowledge graphs} that contain \textbf{relations} between specific \textbf{entities} in triples (for example, ``\textit{Barack Obama}'', ``\textit{isPoliticianOf}'', ``\textit{United States}'') and
(ii) the \textbf{ontology-view knowledge graphs} that constitute semantic \textbf{meta-relations} of abstract \textbf{concepts} (such as ``\textit{polication}'', ``\textit{is leader of}'', ``\textit{city}''). 
In addition, KBs also provide \textbf{cross-view} links that connect ontological concepts and instances, denoting whether an instance is an instantiation from a specific concept. Figure \ref{fig:two-view} shows a snapshot of such a KB. \par

\vspace{-4pt}
In the past decade, KG embedding models have been widely investigated.
These models, which encode KG structures into low-dimensional embedding spaces, are vital to capturing the latent semantic relations of entities and concepts, and support relational inferences in the form of vector algebra.
They have provided an efficient and systematic solution to various knowledge-driven machine learning tasks, such as relation extraction~\cite{wang2014knowledge}, question answering~\cite{bordes2014open}, dialogues agents~\cite{he2017learning},  knowledge alignment~\cite{chen2016multilingual} and visual semantic labeling~\cite{fang2017object}. 
Existing embedding models, however, are limited to only one single view, either on the instance-view graph \cite{bordes2013translating,nickel2016holographic,yang2014embedding} or on the ontology-view graph \cite{chen2018on2vec,gutierrez2018knowledge}. 

Learning to represent a KB 
from both views will no doubt provide more comprehensive insights. On one hand, instance embeddings provide detailed and rich information for their corresponding ontological concepts. For example, by observing many individual musicians, the embedding of its corresponding  concept ``\textit{Musician}'' can be largely determined. On the other hand, a concept embedding provides a high-level summary of its instances, which is extremely helpful when an instance is rarely observed. For example, for a musician who has few relational facts in the instance-view graph, we can still tell his or her rough position in instance embedding space because he or she should not be far away from other musicians.    

\nop{Connecting instance-view graphs with ontology-view graphs for KG embeddings provide a new perspective on how to learn better 
representations for both views of the knowledge.}

\nop{On one hand, the structured information from instance-view graphs are able to be aggregated and transferred to improve the characterization of concepts in ontology-view graphs.
On the other hand, ontological concepts also provide useful categorical information to group related entities.
Besides, a joint learning strategy on both views of a KB 
enables new embedding applications such as entity typing and ontology population. \par
}
\nop{, 
which, unfortunately, still remains unexplored.}

In this paper, we propose to jointly embed the instance-view graph and the ontology-view graph, by leveraging (1) triples in both graphs and (2) cross-view links that connect the two graphs.
It is a non-trivial task to effectively combine representation learning techniques on both views of a KB together, which 
faces the following challenges: 
(1) the vocabularies of entities and concepts, as well as relations and meta-relations, are 
disjoint but semantically related in these two views of the KB, and the semantic mappings from entities to concepts and from relations to meta-relations are complicated and difficult to be precisely captured by any current embedding models;
(2) 
the known cross-view links often inadequately cover a vast number of entities, which leads to insufficient information to align both views of the KB, and curtails discovering new cross-view links;
(3) the scales and topological structures are also largely different
in the two views,  
where the ontological views are often sparser, provide fewer types of relations, and form hierarchical substructures, and the instance view is much larger and with much more relation types. \par

To address the above issues, we propose a novel KG embedding model named \texttt{\modelname}, which jointly encodes 
both the ontology and instance views of a KB.
\nop{
\texttt{\modelname} learns to represent both 
types of model components that learn on the two aspects of the KB. }
\texttt{\modelname} contains two components. First, a \emph{cross-view association model} is designed to associate the instance embedding to its corresponding concept embedding.
Second, the \emph{intra-view embedding model} characterizes the relational facts of ontology and instance views in two separate embedding spaces. 
For the cross-view association model, we explore two techniques to capture the cross-view links. 
The \emph{cross-view grouping} technique assumes that the two views can be forced into the same embedding space, while the \emph{cross-view transformation} technique enables non-linear transformations from the instance embedding space to the ontology embedding space.
\nop{captures the relatedness of concepts and associated entities in embedding spaces of the same dimensionality.
The latter enables non-linear transformations through ontology and instance-specific embedding spaces with different dimensionalities.}
As for the intra-view embedding model, in particular, we use three state-of-the-art translational or similarity-based relational embedding techniques to capture the multi-relational structures of each view.
Additionally, for some KBs where ontologies contain hierarchical substructures, we employ a \emph{hierarchy-aware} embedding technique based on intra-view non-linear transformations to preserve such substructures.
Accordingly, we investigate nine variants of \texttt{\modelname} and evaluate these models on two tasks: the triple completion task and the entity typing task.
Experimental results on the triple completion task
confirm the effectiveness of \texttt{\modelname} for populating knowledge in both ontology and instance-view KGs, which has significantly outperformed various baseline models. 
The results on the entity typing task show that our model is competent in discovering 
cross-view links to align the ontology-view and the instance-view KGs.

\newcommand{\stitle}[1]{\vspace{0.3ex}\noindent{\bf #1}}

\section{Related Work} \label{sec:related}

To the best of our knowledge, there is no previous work on learning to embed two-view knowledge of a KB.
We discuss the following three lines of research work that are closely relevant to this paper.\par

\stitle{Knowledge Graph Embeddings.}
Recent work has put extensive efforts in learning instance-view KG embeddings.
Given triples $(h,r,t)$, where $r$ represents the relation between the head entity $h$ and the tail entity $t$,
the key of KG embeddings is to design a plausibility scoring function $f_r(\mathbf{h},\mathbf{t})$ as the optimization objective ($\mathbf{h}$ and $\mathbf{t}$ are embeddings of $h$ and $t$).
A recent survey~\cite{wang2017knowledge} categorizes the majority of KG embedding models into translational models and similarity-based models.
The representative translational model, TransE~\cite{bordes2013translating}, adopts 
the score function $f_r(\mathbf{h},\mathbf{t}) = -||\mathbf{h} + \mathbf{r} - \mathbf{t}||$ to capture the relation as a translation vector $\mathbf{r}$ between two entity vectors.
Follow-ups of TransE typically vary the translation processes in different forms of relation-specific spaces, so as to improve the performance of triple completion.
Examples include TransH~\cite{wang2014knowledge}, TransR~\cite{lin2015learning}, TransD~\cite{ji2015knowledge} and TransA~\cite{jia2016locally}, etc.
As for the similarity-based models, DistMult~\cite{yang2014embedding} associates related entities using Hadamard product of embeddings, and HolE~\cite{nickel2016holographic} substitutes Hadamard product with circular correlation to 
improve the encoding of asymmetric relations, and achieves the state-of-the-art performance in KG completion.
ComplEx~\cite{trouillon2016complex} migrates DistMult in a complex space 
and offers comparable performance.
Besides, there are other forms of models, including tensor-factorization-based RESCAL~\cite{nickel2011three}, and neural models NTN~\cite{socher2013reasoning} and ConvE~\cite{dettmers2017convolutional}.
These approaches also achieve comparable performances on triple completion tasks at the cost of high model complexity.\par

It is noteworthy that a few approaches have been proposed to incorporate complex type information of entities into above KG embedding techniques \cite{krompass2015type,xie2016representation,ma2017transt,ma2018hierarchical},
from which our settings are substantially different in two perspectives:
(i) These studies utilize the proximity of entity types to strengthen the learning of instance-level entity similarity, while do not capture the semantic relations between such types; 
(ii) They mostly focus on improving instance-view triple completion, but do not leverage instance-view knowledge to improve ontology population, nor support cross-view association to bridge instances and ontological concepts. 
Another related branch on leveraging logic rules~\cite{rocktaschel2015injecting, guo2016jointly,du2017enhancing} requires additional information that typically is not provided in two-view KBs.

\stitle{Multi-graph Embeddings for KGs.}
More recent studies have extended embedding models to bridge multiple KG structures, typically for multilingual learning.
MTransE \cite{chen2016multilingual} thereof, jointly learns a transformation across two separate translational embedding spaces, which can be adopted to our problem.
However, since this multilingual learning approach partly relies on similar structures of KGs, it unsurprisingly falls short of capturing the associations between the
two views of KB with disjoint vocabularies and different topologies, as we show in the experiments.
Later extensions of this model family, such as KDCoE \cite{chen2018co} and JAPE \cite{sun2017cross}, require additional information of literal descriptions \cite{chen2018co} and numerical attributes of entities \cite{sun2017cross} that are typically not available in the ontology views of the KB.
Other models depend on the use of neural machine translation \cite{otani2018cross}, causal reasoning \cite{yeo2018machine} and bootstrapping of strictly 1-to-1 matching of inter-graph entities \cite{zhu2017iterative,sun2018bootstrapping} that do not apply to the nature of our corpora and tasks. 

\stitle{Ontology Population.}
Traditional ontology population is mostly based on extensive manual efforts, or requires large annotated text corpora for the mining of the meta-relation facts
~\cite{wang2006automatic,culotta2004dependency,mousavi2014text,giuliano2008instance}.
These previous approaches rely on intractable parsing or human 
efforts, which generate massive relation facts that are subject to frequent conflicts~\cite{pasternack2013latent}.
A few studies extend embedding techniques to general cross-domain ontologies like ConceptNet.
Examples of such include
On2Vec~\cite{chen2018on2vec} that extends translational embeddings to capture the relational properties and hierarchies of ontological relations,
and 
\citeauthor{gutierrez2018knowledge}~\cite{gutierrez2018knowledge} propose to learn second-order proximity of concepts by combining chained logic rules with ontology embeddings.
This shows the benefits of KG embeddings on predicting relational facts for ontology population, while we argue that such a task can be simultaneously enhanced with the characterization of the instance knowledge.\par 

\section{Modeling}\label{sec:model}

In this section, we introduce our proposed model \texttt{\modelname}, which jointly embed entities and concepts using two model components: \emph{cross-view association model} and \emph{intra-view model}.
We start with the formalization of two-view knowledge bases. \par
\begin{figure}[!ht]
	\centering
	\includegraphics[width=0.9\columnwidth]{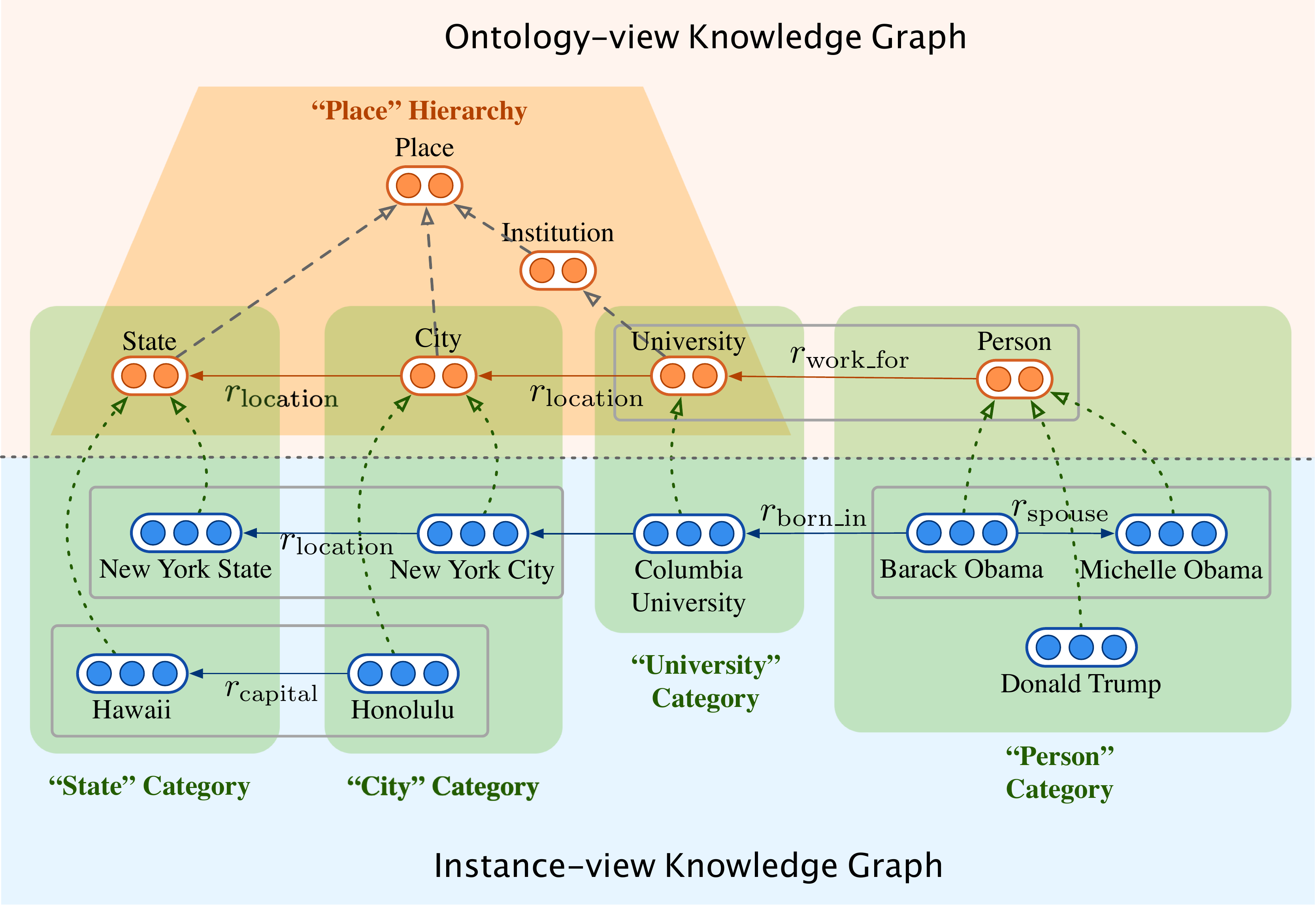}
	\caption{\texttt{\modelname} learns two aspects of a KB. The cross-view association model learns embeddings from cross-view links (dash arrows in green ``category'' box). The default intra-view model learns embeddings from triples (grey box) in each view; Besides, hierarchy-aware intra-view models the meta-relation facts that form hierarchies in the ontology (orange   ``Hierarchy'' trapezoid).}
	\label{fig:model_archi}
\end{figure}
\vspace{-12pt}
\subsection{Formalization of Knowledge Bases}\label{subsec:formulation}
In a KB, we use $\mathcal{G}_I$ and $\mathcal{G}_O$ to denote the instance-view KG and ontology-view KG respectively. 
The instance-view KG is denoted as $\mathcal{G}_I$, which is formed with $\mathcal{E}$, the set of entities, and $\mathcal{R}_I$, the set of relations. 
The set of concepts and meta-relations in the ontology-view graph $\mathcal{G}_O$ are similarly denoted as $\mathcal{C}$ and $\mathcal{R}_O$ respectively. Note that $\mathcal{E}$ and $\mathcal{C}$ (or $\mathcal{R}_I$ and $\mathcal{R}_O$) are disjoint sets.
$(h^{(I)},r^{(I)},t^{(I)})\in\mathcal{G}_I$ and $(h^{(O)},r^{(O)},t^{(O)})\in\mathcal{G}_O$ denote triples in the instance-view KG and the ontology-view KG respectively, such that $h^{(I)}, t^{(I)} \in \mathcal{E}, h^{(O)}, t^{(O)} \in \mathcal{C}$, $r^{(I)} \in \mathcal{R}_I$, and $r^{(O)} \in \mathcal{R}_O$.
Specifically, for each view in the KB, a dedicated low-dimensional space is assigned to embed nodes and edges. 
Boldfaced $\mathbf{h}^{(I)},\mathbf{t}^{(I)},\mathbf{r}^{(I)}$  represent the embedding vectors of head entity $h^{(I)}$, tail entity $t^{(I)}$ and relation $r^{(I)}$ in instance-view triples. Similarly,  $\mathbf{h}^{(O)},\mathbf{t}^{(O)}$, and  $\mathbf{r}^{(O)}$ denote the embedding vectors for the corresponding concepts and their meta-relation in the ontology-view graph.
Besides the notations for two views, $\mathcal{S}$ is used to denote the set of known cross-view links in the KB, which contains associations between instances and concepts such as \textit{``type\_of''}. 
We use $(e,c) \in \mathcal{S}$ to denote a link between $e \in \mathcal{E}$ and its corresponding concept $c \in \mathcal{C}$.
\nop{
is one sample of such links where $c$ in the pair always refers to a concept and $e$ is one real-world instantiation of 
$c$. }
For example, ($e$: Los Angeles International Airport, $c$: airport) denotes that ``\textit{Los Angeles International Airport}'' is an instance of the concept ``\textit{airport}''.
Looking into the nature of the ontology view, we also have hierarchical substructures identified by \textit{``subclass\_of''} (or other similar meta-relations). That is, we can observe concept pairs $(c_l, c_h) \in \mathcal{T}$ that indicates a finer (more specific) concept belongs to a coarser (more general) concept. One aforementioned example is ($c_l$: singer, $c_h$: person).\par

Our model \texttt{\modelname} consists of two model components that learn embeddings from the two views: the cross-view association model enables the connection and information flow between the two views by capturing the instantiation of entities from corresponding concepts, and the intra-view model encodes the entities/concepts and relations/meta-relations on each view of the KB. 
The illustration of these model components for learning different aspects of the KB is shown in Figure~\ref{fig:model_archi}.
In the following subsections, we first discuss the cross-view association model and intra-view model for each view, then combine them into variants of proposed \texttt{\modelname}\ model.
\vspace{-3pt}
\subsection{Cross-view Association Model} \label{subsec:cat_model}
The goal of the cross-view association model is to capture the associations between the entity embedding space and the concept embedding space, based on the cross-view links in KBs, which will be our key contributions.
We propose two techniques to model such associations: \emph{Cross-view Grouping (CG)} and \emph{Cross-view Transformation (CT)}. These two techniques are based on different assumptions and thus optimize different objective functions.

\begin{figure}
\centering
\begin{subfigure}[b]{1\columnwidth}
   \centering
   \includegraphics[width=0.8\columnwidth]{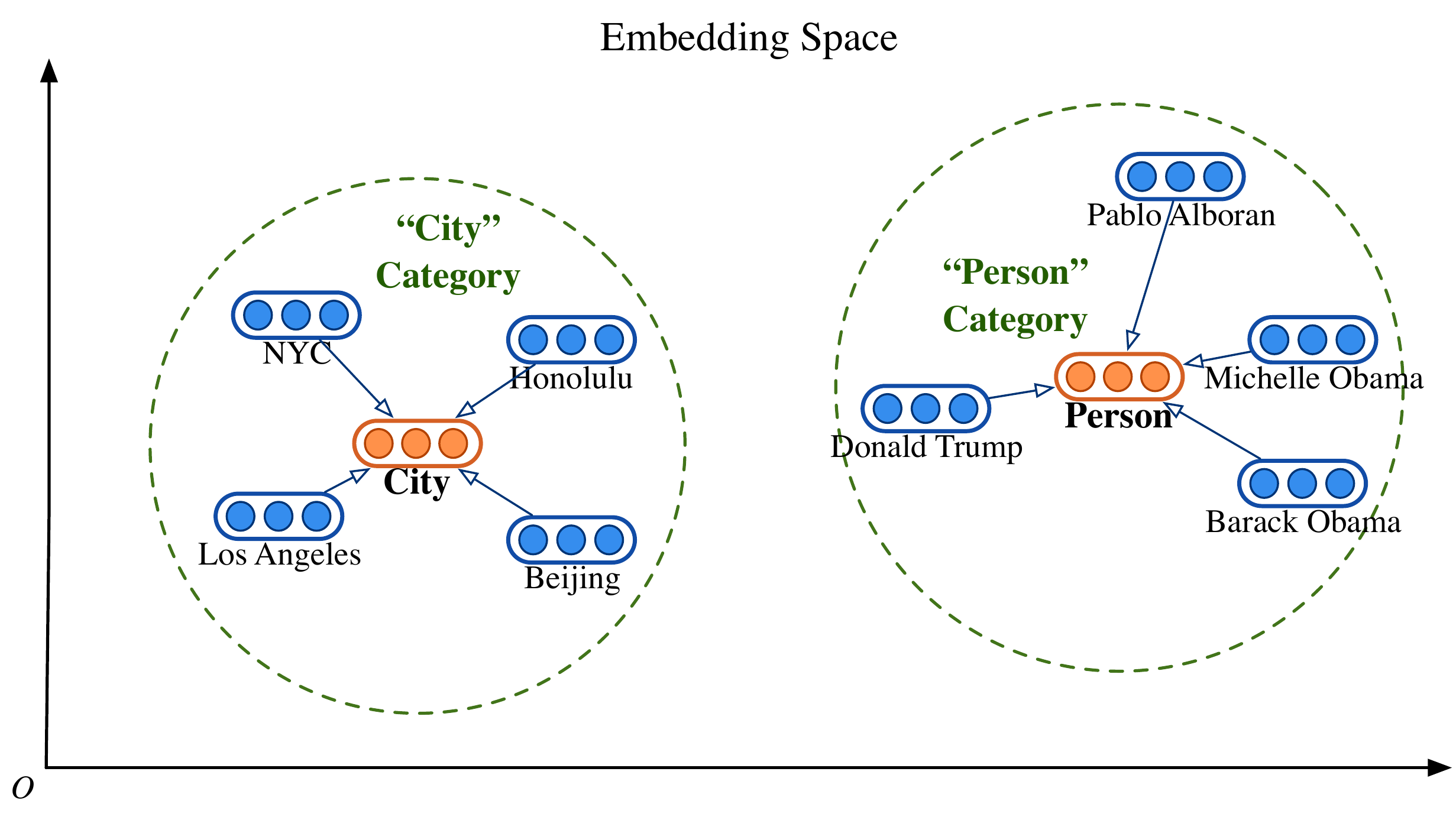}
   \caption{Cross-view Grouping (CG)}
   \label{fig:CG} 
\end{subfigure}

\begin{subfigure}[b]{1\columnwidth}
   \centering
   \includegraphics[width=0.8\columnwidth]{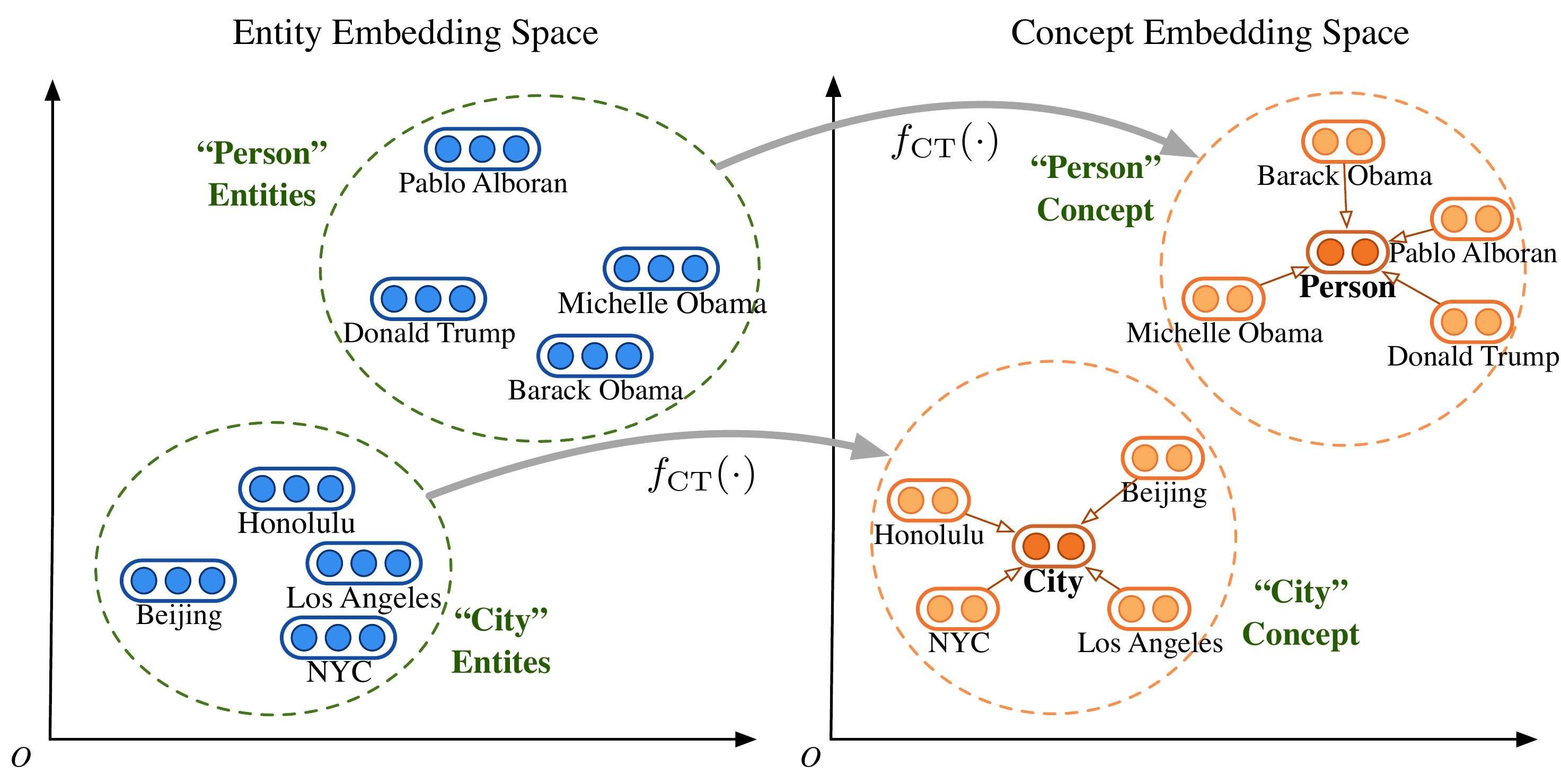}
   \caption{Cross-view Transformation (CT)}
   \label{fig:CT}
\end{subfigure}
\caption{Intuition of the cross-view association model: Cross-view Grouping (a); Cross-view Transformation (b).}
\end{figure}

\stitle{Cross-view Grouping (CG).} 
The cross-view grouping method can be considered as grouping-based regularization, which assumes that the ontology-view KG and instance-view KG can be embedded into the same space, and forces any instance $e\in \mathcal{E}$ to be close to its corresponding concept $c\in \mathcal{C}$, as shown in Figure \ref{fig:CG}.
This requires the embedding dimensionalities for the instance-view and ontology-view graphs to be the same, i.e. $d=d_c=d_e$. 
Specifically, the categorical association loss for a given pair of cross-view link $(e, c)$ is defined as the distance between the embeddings of $e$ and $c$ compared with margin $\gamma^{\text{CG}}$,  and the loss is defined as,
\begin{equation}
\begin{aligned} 
\label{equ:asso_loss_2}
	J_{\text{Cross}}^{\mathrm{CG}}  = \frac{1}{|\mathcal{S}|}\sum_{(e,c)\in\mathcal{S}} \left[ ||\mathbf{c} - \mathbf{e}||_2 - \gamma^{\mathrm{CG}} \right]_{+},
\end{aligned}
\end{equation}
where $[x]_{+}$ is the positive part of the input $x$, i.e. $[x]_{+} = \max\{x,0\}$. This penalizes the case where the embedding of $e$ falls out the $\gamma^{\text{CG}}$-radius\footnote{Typically, margin hyperparameter $\gamma$ in the hinge loss can be chosen as 0.5 or 1 for different model settings. However, it is not a sensitive hyperparameter in our models.} neighborhood centered at the embedding of $c$. 
CG has a strong clustering effect 
that makes entity embeddings close to their concept embeddings in the end. 

\stitle{Cross-view Transformation (CT).}
We also propose a cross-view transformation technique, which seeks to transform information between the entity embedding space and the concept space. Unlike CG that requires the two views to be embedded into the same space, the CT technique allows the two embedding spaces to be completely different from each other, which will be aligned together via a transformation, as shown in Figure \ref{fig:CT}.
\nop{
CT assumes that the entity embeddings shall be close to the target concept embedding after their aggregation through entity-concept transformation function, i.e.,}
In other words, after the transformation, an instance will be mapped to an embedding in the ontology-view space, which should be close to the embedding of its corresponding concept:
\begin{equation}
	\mathbf{c} \leftarrow f_{\text{CT}} \left(\mathbf{e} \right),  \forall (e,c)\in\mathcal{S},
	\label{equ:GCN}
\end{equation}
where $f_{\text{CT}}(\mathbf{e})=\sigma(\mathbf{W_{\text{ct}}}\cdot\mathbf{e} + \mathbf{b}_\text{ct})$ is a non-linear affine transformation. 
$\mathbf{W_{\text{ct}}} \in \mathbb{R}^{d_2 \times d_1}$ thereof is a weight matrix and $\mathbf{b}_{\text{ct}}$ is a bias vector. 
$\sigma(\cdot)$ is a non-linear activation function, for which we adopt $\tanh$.

Therefore, the total loss of the cross-view association model is formulated as Equation \ref{equ:asso_loss_3}, which aggregates the CT objectives for all concepts involved in $\mathcal{S}$.

\begin{equation}
\begin{aligned} 
\label{equ:asso_loss_3}
J_{\text{Cross}}^{\mathrm{CT}} = \frac{1}{|\mathcal{S}|} \sum_{\substack{(e,c)\in\mathcal{S} \\ \wedge (e,c') \notin\mathcal{S}}}\left [\gamma^{\text{CT}} + \left|\left| \mathbf{c} - f_{\text{CT}}(\mathbf{e}) \right|\right|_2
- \left|\left| \mathbf{c}' - f_{\text{CT}}(\mathbf{e}) \right|\right|_2 \right ]_+
\end{aligned}
\end{equation}

\vspace{-5pt}
\subsection{Intra-view Model} \label{subsec:struct_model}

The aim of intra-view model is to preserve the original structural information in each view of the KB separately in two embedding spaces. Because of the different semantic meanings of relations in the instance view and meta-relations in the ontology view, it helps to give each view separate treatment rather than combining them into a single representation schema, improving the performance of downstream tasks, as shown in Section \ref{subsec:ex_link}. 
In this section, we provide two intra-view model techniques for encoding heterogeneous and hierarchical graph structures.

\stitle{Default Intra-view Model.}
To embed such a triple $(h,r,t)$ in one KG, a score function $f(\mathbf{h},\mathbf{r},\mathbf{t})$ measures the plausibility of it. 
A higher score indicates a more plausible triple.
Any triple embedding technique is applicable in our intra-view framework. In this paper, we adopt three representative techniques, i.e. translations~\cite{bordes2013translating}, multiplications~\cite{yang2014embedding} and circular correlation~\cite{nickel2016holographic}. The score functions of these techniques are given as follows.
\begin{equation}
	\begin{aligned}
		f_{\mathrm{TransE}}(\mathbf{h},\mathbf{r},\mathbf{t}) & = -||\mathbf{h} + \mathbf{r} - \mathbf{t}||_2\\
		f_{\mathrm{Mult}}(\mathbf{h},\mathbf{r},\mathbf{t}) & = (\mathbf{h}\circ\mathbf{t})\cdot\mathbf{r}\\
		f_{\mathrm{HolE}}(\mathbf{h},\mathbf{r},\mathbf{t})  & = (\mathbf{h} \star \mathbf{t})\cdot\mathbf{r}
	\label{equ:triple_loss}
	\end{aligned}
\end{equation}
where $\circ$ is the Hadamard product and $\cdot$ is the dot product. $\star:\mathbb{R}^{d}\times\mathbb{R}^{d}\rightarrow\mathbb{R}^{d}$ denotes circular correlation defined as $[\mathbf{a} \star \mathbf{b}]_k = \sum^{d}_{i=0} a_{i}b_{(k+i)\mod d}$.

To learn embeddings of all nodes in one graph $\mathcal{G}$, a hinge loss is minimized for all triples in the graph:
\begin{equation}
	\label{equ:margin_loss}
	J_{\text{Intra}}^{\mathcal{G}} = \frac{1}{|\mathcal{G}|}\sum_{\substack{ (h,r,t) \in \mathcal{G}  \\ \wedge (h',r,t') \notin \mathcal{G}} }\left[\gamma^{\mathcal{G}}+f(\mathbf{h',r,t'})-f(\mathbf{h,r,t}) \right]_{+},
\end{equation}
where $\gamma^{\mathcal{G}}>0$ 
is a  positive margin, and $(h',r,t')$ is one sample from the set of corrupted triples which replace either head or tail entity and does not exist in $\mathcal{G}$. 

The aforementioned techniques, losses and learning objectives for embedding graphs are naturally applicable for both  instance-view graph and  ontology-view graph. 
In the default intra-view model setting, for triples $(h^{(I)},r^{(I)},t^{(I)}) \in \mathcal{G}_I$ or $(h^{(O)},r^{(O)},t^{(O)}) \in \mathcal{G}_O$, we can compute $f_I(\mathbf{h}^{(I)},\mathbf{r}^{(I)},\mathbf{t}^{(I)})$ and $f_O(\mathbf{h}^{(O)},\mathbf{r}^{(O)},\mathbf{t}^{(O)})$ with the same techniques when optimizing $J_{\text{Intra}}^{\mathcal{G}_I}$ and $J_{\text{Intra}}^{\mathcal{G}_O}$. 
Combining the loss from instance-view and ontology-view graphs, the joint loss of the intra-view model is given as below,
\begin{equation} 
\label{equ:struct_loss}
	J_{\text{Intra}} = J_{\text{Intra}}^{\mathcal{G}_I} + \alpha_1  \cdot \ J_{\text{Intra}}^{\mathcal{G}_O},
\end{equation}
where a positive hyperparameter $\alpha_1$ 
weighs between the structural loss of the instance-view graph and ontology-view graph. 
\nop{Also, $|\mathcal{G}_I|$ and $|\mathcal{G}_O|$ denote the number of triples in $\mathcal{G}_I$ and $\mathcal{G}_O$, which indicates that the structure loss from each graph is averaged by the number of triples in the corresponding view.}

In \texttt{\modelname} deployed with the default Intra-view model, we employ the same triple encoding technique to represent both views of the KB.
The purpose of doing so is to enforce the same paradigm of characterizing relational inferences in both views. 
It is noteworthy that there are other triple encoding techniques for KG embeddings, which can potentially be used in our intra-view model. 
Since exploring different triple encoding techniques is not the focus of our paper, we leave them as future work.
\nop{Due to the high parameter complexity (TransA, ConVe, etc) or embedding on non-euclidean space (ComplEx, etc), we currently do not consider them as options for the intra-view model for \texttt{\modelname}.}

\stitle{Hierarchy-Aware Intra-view Model for the Ontology.} 
It is observed that the ontology view of some KBs form hierarchies, which is typically constituted by a meta-relation with the hierarchical property, such as  ``\textit{subclass\_of}'' and ``\textit{is\_a}''~\cite{mahdisoltani2014yago3,lehmann2015dbpedia}. We can define such meta-relation facts as ($c_l$, $r_\text{meta}$ = ``\textit{subclass\_of}'', $c_h$). 
For example, ``\textit{musician}'' and ``\textit{singer}'' belong to ``\textit{artist}'' and ``\textit{artist}'' is also subclass of ``\textit{person}''.
Such semantic ontological features requires additional modeling than other meta-relations. In other words, we further distinguish between meta-relations that form the ontology hierarchy and those regular semantic relations (such as \textit{``related\_to''}) in our intra-view model.

To address this problem, we propose the hierarchy-aware (HA) intra-view model by extending a similar method to that of cross-view transformation as defined in Equation~\ref{equ:GCN}. Given concept pairs $(c_l,c_h)$, we model such hierarchies into a non-linear transformation between coarser concepts and associated finer concepts by
\begin{equation}
	g_{\text{HA}}(\mathbf{c}_h)=\sigma(\mathbf{W}_{\text{HA}}\cdot\mathbf{c}_l + \mathbf{b}_{\text{HA}})
	\label{equ:Onto-GCN}
\end{equation}
where $\mathbf{W}_{\text{HA}} \in \mathbb{R}^{d_2 \times d_2}$ and $\mathbf{b}_{\text{HA}} \in \mathbb{R}^{d_2}$ are defined similarly. Also, we use $\tanh$ function as $\sigma(\cdot)$ option. This will introduce a new loss term, ontology hierarchy loss inside the ontology view, which is similar to Equation \ref{equ:asso_loss_3}, 
\begin{equation}
J_{\text{Intra}}^{\mathrm{HA}} = \frac{1}{|\mathcal{T}|} \sum_{\substack{(c_l,c_h)\in\mathcal{T} \\ \wedge (c_l,c'_h) \notin\mathcal{T}}} \left[\gamma^{\text{HA}} + \left|\left| \mathbf{c}_h - g(\mathbf{c}_l) \right|\right|_2
- \left|\left| \mathbf{c_h}' - g(\mathbf{c_l}) \right|\right|_2 \right]_+
\end{equation}
Therefore, the total training loss of the hierarchy-aware intra-view model for both views changes slightly to,
\begin{equation}
\begin{aligned}
	J_{\text{Intra}} &= J_{\text{Intra}}^{\mathcal{G}_I} + \alpha_1 \cdot J_{\text{Intra}}^{\mathcal{G}_O \backslash \mathcal{T}} + \alpha_2 \cdot J_{\text{Intra}}^{\mathrm{HA}}
	\label{equ:intra-ha}
\end{aligned}
\end{equation}
where positive $\alpha_1$ and $\alpha_2$ are two weighing hyperparameters. In Equation \ref{equ:intra-ha}, $J_{\text{Intra}}^{\mathcal{G}_O \backslash \mathcal{T}}$ refers to the loss of the default intra-view model that is only trained on triples with regular semantic relations.  $J_{\text{Intra}}^{\mathrm{HA}}$ is explicitly trained on the triples with meta-relations that form the ontology hierarchy, which is a major difference from Equation \ref{equ:struct_loss}.

As the conclusion of this subsection, in \texttt{\modelname}, the basic assumption is that KGs have ontology hierarchy and rich semantic relational features compared to social or citation networks.  \texttt{\modelname} is able to encode such KG properties in its model architecture. Note that we are also aware of the fact that there are more comprehensive properties of relations and meta-relations in the two views such as logical rules of relations and entity types. Incorporating such properties into the learning process is left as future work.

\subsection{Joint Training on Two-View KBs}

Combining the intra-view model and cross-view association model, \texttt{\modelname} minimizes the following joint loss function:
\begin{equation}
\begin{aligned} 
\label{equ:total_loss}
	J = J_{\text{Intra}} + \omega \cdot J_{\text{Cross}},
\end{aligned}
\end{equation}
where  $\omega>0$ is positive hyperparameter that balances between $J_{\text{Intra}}$ and $J_{\text{Cross}}$.

Instead of directly updating $J$, our implementation optimizes $J_{\text{Intra}}^{\mathcal{G}_I}$, $J_{\text{Intra}}^{\mathcal{G}_O} $ and $ J_{\text{Cross}}$ alternately. 
In detail, we optimize $\theta^{\text{new}} \leftarrow \theta^{\text{old}} -\eta\nabla J_{\text{Intra}}$ and $\theta^{\text{new}} \leftarrow \theta^{\text{old}} - (\omega\eta)\nabla J_{\text{Cross}}$ 
in successive steps within one epoch. $\eta$ is the learning rate, and $\omega$ differentiates between the learning rates for intra-view and cross-view losses. 

We use the AMSGrad optimizer~\cite{reddi2018convergence} to optimize the joint loss function. 
We initialize vectors by drawing from a uniform distribution on the unit spherical surface, and initialize matrices using random orthogonal initialization~\cite{saxe2013exact}. 
During the training, we enforce the constraint that the L2 norm of all entity and concept vectors to be 1, in order to prevent them from shrinking to zero. 
This follows the setting by \cite{bordes2013translating, yang2014embedding, wang2014knowledge, nickel2016holographic}.
Negative sampling is used on both intra-view model and cross-view association model with a ratio of 1 (number of negative samples per positive one).
A hinge loss is applied for both models with all variants.

\subsection{Variants of \texttt{\modelname} and Complexity} 

Without considering the HA technique, we have six variants of \texttt{\modelname} given two options of cross-view association models in Section \ref{subsec:cat_model} and three options of intra-view models in Section \ref{subsec:struct_model}.
For simplicity, we use the names of its components to denote specific variants of \texttt{\modelname}, such as ``\texttt{\modelname}-TransE-CT'' represents \texttt{\modelname} with the cross-view transformation and TransE-based default intra-view embeddings.
In addition, we incorporate the hierarchy-aware intra-view model 
for the ontology view into cross-view transformation model\footnote{We later show in the experiments that CT-based variants consistently outperform CG-based variants and thus we only apply HA intra-view model settings to CT-based model variants.}, which produces three additional model variants denoted as {\texttt{\modelname}-HATransE-CT}, {\texttt{\modelname}-HAMult-CT}, and {\texttt{\modelname}-HAHolE-CT}. 

The model complexity depends on the cross-view association model and intra-view model for learning two-view KBs. 
We denote $n_e,n_c,n_r,n_m$ as the number of total entities, concepts, relations and meta-relations (typically $n_e\gg n_c$) and $d_e, d_c$ as embedding dimensions ($d_e=d_c$ if CG is used).
The model complexity of parameter sizes is $\mathcal{O}(n_ed_e+n_cd_c)$ for all CG-based variants and $\mathcal{O}(n_ed_e+n_cd_c+d_ed_c)$ for all CT-based variants. 
An additional parameter size of $\mathcal{O}(d^2_c)$ is needed if the hierarchy-aware intra-view model applies. Because of $n \gg d_e$ (or $d_c$), the parameter complexity is approximately proportional to the number of entities and the model training runtime complexity is proportional to the number of triples in the KG.
For the task of triple completion in the KG, the time complexity for all variants is $\mathcal{O}(n_ed_e)$ for the instance-view graph or $\mathcal{O}(n_cd_c)$ for the ontology-view graph. To process each prediction case in the entity typing task, the time complexity is $\mathcal{O}(n_cd_e)$ for CG and $\mathcal{O}(n_cd_cd_e)$ for CT.
Details about each task are curated in Section \ref{subsec:ex_link} and \ref{subsec:ex_type}.
\def\YAGO{YAGO26K-906}
\def\DBpedia{DB111K-174}

\section{Experiments}\label{sec:experiment}
In this section, we evaluate \texttt{\modelname}  with two groups of tasks: the triple completion task (Section \ref{subsec:ex_link}) on both instance-view and ontology-view KGs and the entity typing task (Section \ref{subsec:ex_type}) to bridge two views of the KB. 
Besides, we provide a case study in Section \ref{subsec:case} on ontology population and long-tail entity typing. We also present hyperparameter study, effects of cross-view sufficiency and negative samples in Appendix \ref{sec:sup-discuss}.


\subsection{Datasets} 
To the best of our knowledge, existing datasets for KG embeddings consider only an instance view (e.g. FB15k \cite{bordes2013translating}) or an ontology view (e.g. WN18 \cite{bordes2014semantic}).
Hence, we prepare two new datasets: \emph{YAGO26K-906} and \emph{DB111K-174}, which are extracted from YAGO~\cite{mahdisoltani2014yago3} and DBpedia~\cite{lehmann2015dbpedia} respectively. The detailed dataset construction process is described in Appendix \ref{sec:sup-dataset}.

Table \ref{tab:dataset} provides the statistics of both datasets. Normally, the instance-view KG is significantly larger than the ontology-view graph. Also, we notice that the two KBs are different in the density of type links,  
i.e., \DBpedia\ has a much higher entity-to-concept ratio (643.4) than \YAGO\ (28.7).
Datasets are available at \href{https://github.com/JunhengH/joie-kdd19}{\texttt{https://github.com/JunhengH/joie-kdd19}}.

\begin{table*}
\centering
\caption{Statistics of datasets.}
\vspace{-4pt}
{
\small
\begin{tabular}{c|ccc|ccc|cc}
\newtoprule
\newtoprule
\multirow{2}{*}{Dataset} &
\multicolumn{3}{c|}{Instance Graph $\mathcal{G}_I$} &
\multicolumn{3}{c|}{Ontology Graph $\mathcal{G}_O$} &\multirow{2}{*}{Type Links $\mathcal{S}$} \\
  & \#Entities & \#Relations & \#Triples & \#Concepts & \#Meta-relations & \#Triples & {} \\
\newmidrule
\YAGO & 26,078 & 34 & 390,738 & 906 & 30 & 8,962 & 9,962  \\
\DBpedia & 111,762 & 305 & 863,643 & 174 & 20 & 763 & 99,748  \\
\newbottomrule
\newbottomrule
\end{tabular}
}
\label{tab:dataset}
\end{table*}
\vspace{-5pt}

\subsection{KG Triple Completion}\label{subsec:ex_link} 

The objective of triple completion is to construct the missing relation facts in a KG structure, which directly tests the quality of learned embeddings. 
In our experiment, this task spans into two sub-tasks for instance-view KG completion and ontology population. We perform the sub-tasks on both datasets with all \texttt{\modelname} variants compared with baseline models. 

\stitle{Evaluation Protocol}
First, we separate the instance-view triples into training set $\mathcal{G}_I^{\text{train}}$, validation set $\mathcal{G}_I^{\text{valid}}$ and test set $\mathcal{G}_I^{\text{test}}$, as well as separate similarly the ontology-view triples to $\mathcal{G}_O^{\text{train}}$, $\mathcal{G}_O^{\text{valid}}$ and $\mathcal{G}_O^{\text{test}}$. The percentage of the training, validation and  test cases is approximately 85\%, 5\% and 10\%, which is consistent to that of the widely used benchmark dataset~\cite{bordes2013translating} for instance-only KG embeddings.
Each \texttt{\modelname} variant is trained on $\mathcal{G}_I^{\text{train}}$ and $\mathcal{G}_O^{\text{train}}$ triples along with all cross-view links $\mathcal{S}$. 
In the testing phase, given each query $(h,r,?t)$, the plausibility scores $f(\mathbf{h},\mathbf{r},\mathbf{\tilde{t}})$ for triples formed with every $\tilde{t}$ in the test candidate set are computed and ranked by the intra-view model. 
We report three metrics for testing: mean reciprocal ranks ($MRR$), accuracy ($Hits@1$) and the proportion of correct answers ranked within the top 10 ($Hits@10$).
All three metrics are preferred to be higher, so as to indicate better triple completion performance. 
Also, we adopt the filtered metrics as suggested in previous work which are aggregated based on the premise that the candidate space has excluded the triples that have been seen in the training set~\cite{bordes2013translating,yang2014embedding}.

As for the hyperparameters in training, we select the dimensionality $d$ among $\{50, 100, 200, 300\}$ for concepts and entities, learning rate among $\{0.0005,0.001,0.01\}$, 
margin $\gamma$ among $\{0.5, 1\}$. We also use different batch sizes according to the sizes of graphs. We fix the best configuration $d_e=300, d_c=50$ for CT and $d_e=d_c=200$ for CG with $\alpha_1=2.5, \alpha_2=1.0$. 
We set $\gamma^{\mathcal{G}_I}=\gamma^{\mathcal{G}_O}=0.5$ as the default for all TransE variants and $\gamma^{\mathcal{G}_I}=\gamma^{\mathcal{G}_O}=1$ for all Mult and HolE variants. The training processes on all datasets and models are limited to 120 epochs.

\stitle{Baselines} 
We compare our model with TransE, DistMult and HolE as well as TransC~\cite{lv2018differentiating}. 
We deploy the following variants of baselines:
(i) We train these mono-graph models (TransE, DistMult and HolE) either on instance-view triples or ontology-view triples separately, denoted as (\emph{base}) in Table~\ref{tab:link};  
(ii) We also train TransE, DistMult and HolE based on all triples in both $\mathcal{G}_I^{\text{train}}$ and $\mathcal{G}_O^{\text{train}}$. For the second setting thereof, we incorporate cross-view links by adding one additional relation ``\textit{type\_of}'' to them, denoted as (\emph{all}) in Table~\ref{tab:link}. 
(iii) 
TransC is trained on both views of a KB.
TransC is a recent work that differentiates between the encoding process of concepts from instances. Note that TransC is equivalent to a simplified case of our \texttt{\modelname}-TransE-CG where no semantic meta relations in the ontology view are included.
For that reason, TransC does not apply to the completion of the ontology view.
\begin{table*}[!htb]
\caption{Results of KG triple completion. H@1 and H@10 denote $Hit@1$ and $Hit@10$ respectively. For each group of model variants with the same intra-view model, the best results are bold-faced. The overall best results on each dataset are underscored.} 
\centering
{
\small
\begin{tabular}{c|ccc|ccc|ccc|ccc}
\newtoprule
\newtoprule
{Datasets} &
\multicolumn{6}{c|}{\YAGO} &
\multicolumn{6}{c}{\DBpedia}
\\
\newmidrule
{Graphs} &
\multicolumn{3}{c|}{$\mathcal{G}_I$ KG Completion} &
\multicolumn{3}{c|}{$\mathcal{G}_O$ KG Completion} & 
\multicolumn{3}{c|}{$\mathcal{G}_I$ KG Completion} &
\multicolumn{3}{c}{$\mathcal{G}_O$ KG Completion}
\\
\newmidrule
{Metrics} & 
MRR & H@1 & H@10 &
MRR & H@1 & H@10 &
MRR & H@1 & H@10 &
MRR & H@1 & H@10 
\\
\newtoprule
\newtoprule
{TransE (base)} & 0.195 & 14.09 & 34.51 & 0.145 & 12.29 & 20.59 & 0.327 & 22.26 & 49.01 & 0.313 & 23.22 & 46.91\\
{TransE (all)} & 0.187 & 13.73 & 35.05 & 0.189 & {14.72} & 24.36 & 0.318 & 22.70 & 48.12 & 0.539 & 47.90 & 61.84 \\
TransC & 0.252 & 15.71 & 37.79 & -- & -- & -- & 0.359 & 24.83 & 49.31 & -- & -- & --\\
\rowcolor{shadecolor} {\texttt{\modelname}-TransE-CG} & 0.264 & 16.38 & 35.45 & 0.189 & 11.16 & 29.44 & 0.394 & 27.75 & 51.20 & 0.598 & 53.84 & 71.79\\
\rowcolor{shadecolor} {\texttt{\modelname}-TransE-CT} & {0.292} & \textbf{18.72} & {44.14} & {{0.240}} & 14.49 & {{33.47}} & {{0.443}} & {{32.10}} & {{67.89}} & \underline{\textbf{0.622}} & \underline{\textbf{58.10}} & {{72.97}}\\
\rowcolor{shadecolor} {\texttt{\modelname}-HATransE-CT} & \textbf{0.306} & 18.62 & \textbf{51.72} & \underline{\textbf{0.263}} & \textbf{16.72} & \underline{\textbf{38.46}} & \underline{\textbf{0.473}} & \underline{\textbf{33.79}} & \underline{\textbf{71.37}} & 0.591 & 52.07 & \underline{\textbf{79.65}}\\
\newmidrule
{DistMult (base)} & 0.253 & 22.91 & 28.76 & 0.197 & \textbf{17.72} & 25.08 & 0.265 & 25.95 & 27.63 & 0.235 & 15.18 & 29.11\\
{DistMult (all)} & 0.288 & \textbf{24.06} & 31.24 & 0.156 & 14.32 & 16.54 & 0.280 & {27.24} & 29.70 & 0.501 & 45.52 & 64.73\\ 
\rowcolor{shadecolor} {\texttt{\modelname}-Mult-CG} & 0.274 & 18.80 & 37.45 & 0.198 & 11.16 & 27.91 & 0.320 & 23.44 & 49.49 & 0.532 & 46.15 & 68.91\\ 
\rowcolor{shadecolor} {\texttt{\modelname}-Mult-CT} & \textbf{0.309} & 20.40 & \textbf{46.15} & \textbf{0.207} & 14.71 & 30.43 & \textbf{0.404} & \textbf{26.55} & \textbf{60.86} & \textbf{0.563} & \textbf{50.50} & {71.62}\\
\rowcolor{shadecolor} {\texttt{\modelname}-HAMult-CT} & 0.296 & 19.39 & 45.48 & 0.202 & 13.72 & \textbf{31.10} & 0.369 & 24.82 & 55.86 & 0.521 & 38.46 & \textbf{77.25}\\
\newmidrule
{HolE (base)} & 0.265 & \underline{\textbf{25.90}} & 28.31 & 0.192 & 18.70 & 20.29 & 0.301 & {29.24} & 31.51 & 0.227 & 18.91 & 32.83\\
{HolE (all)} & 0.252 & 24.22  & 26.56 & 0.138 & 11.29 & 14.43 & 0.295 & 28.70 & 30.32 & 0.432 & 38.80 & 56.05\\
\rowcolor{shadecolor} {\texttt{\modelname}-HolE-CG} & 0.253 & 18.75 & 34.11 & 0.167 & 13.04 & 22.33 & 0.361 & 24.13 & 46.15 & 0.469 & 41.89 & 62.16\\
\rowcolor{shadecolor} {\texttt{\modelname}-HolE-CT} & {{0.313}} & 20.40 & {47.80} & 0.229 & \underline{\textbf{20.85}} & {28.42} & {0.425} & 29.09 & {66.88} & \textbf{0.514} & \textbf{43.24} & {69.23}\\
\rowcolor{shadecolor} {\texttt{\modelname}-HAHolE-CT} & \underline{\textbf{0.327}} & 22.42 & \underline{\textbf{52.41}} & \textbf{0.236} & 16.72 & \textbf{30.96} & \textbf{0.464} & \textbf{33.11} & \textbf{69.56} & 0.503 & 40.80 & \textbf{71.03}\\

\newbottomrule
\newbottomrule
\end{tabular}
}
\label{tab:link}

\end{table*}

\stitle{Results} 
As reported in Table \ref{tab:link}, we categorize the results into three different groups based on the intra-view models. 
Though three intra-view models have different capabilities, among all the baselines in same group, \texttt{\modelname} notably outperforms others by 6.8\% on $MRR$, and 14.8\% on $Hit@10$ on average.
A significant improvement is achieved on the ontology-view of \DBpedia\ with \texttt{\modelname} compared to concept embeddings trained with only ontology-view triples and even 10.4\% average increment compared to ``all''-setting baselines and 34.97\% compared to ``base''-setting baselines. These results indicate that \texttt{\modelname} has better ability to utilize information from the instance view to promote the triple completion in ontology view.
Comparing different intra-view models, translation based models 
performs better than similarity based models on ontology population and instance-view KG completion on the \DBpedia dataset. 
This is because these graphs are sparse, and TransE is less hampered by the sparsity in comparison to the similarity-based techniques~\cite{pujara2017sparsity}.
By applying the HA technique in the intra-view models with CT, the performance on instance-view triple completion is noticeably improved in most cases in comparison to the default intra-view CT-based models, especially in variants with translation and circular correlation based intra-view models. 

Generally, \texttt{\modelname} provides an effective method to train two-view KB separately and both $\mathcal{G}_I$ and $\mathcal{G}_O$ benefit each other in learning better embeddings, producing promising results in the triple completion task. 
\vspace{-10pt}
\subsection{Entity Typing}\label{subsec:ex_type}

The entity typing task seeks to predict the associating concepts of certain given entities. Similar to the triple completion task, we rank all candidates and report the top-ranked answers for evaluation.

\stitle{Evaluation Protocol}
We separate the cross-view links of each dataset into training and test sets with the ratio of 60\% to 40\%, denoted as $\mathcal{S}^{\text{train}}$ and $\mathcal{S}^{\text{test}}$ respectively.
Each model is trained on the entire instance-view and ontology-view graphs with cross-view links $\mathcal{S}^{\text{train}}$. 
Hyperparameters are carried forward from the triple completion task, in order to evaluate under controlled variables.
In the test phase, given a specific entity $e_q$, 
we rank the concepts based on their embedding distances from the projection of $\mathbf{e}_q$ in the concept embedding space.
 and calculate $MRR$, $Hit@1$ (i.e. accuracy) and $Hit@3$ on the test queries. We perform the entity typing task on both datasets with all \texttt{\modelname} variants compared with these baselines.

\stitle{Baselines} 
We compare with TransE, DistMult, HolE and MTransE. 
For baselines other than MTransE, we convert the cross-view links $(e,c)$ to triples ($e$, $r_T$=\textit{``type\_of''}, $c$).
Therefore, entity typing is equivalent to the triple completion task for these baseline models.
For MTransE, we treat concepts and entities as different views (originally input as knowledge bases of two languages in \cite{chen2016multilingual}) in their model and test with distance-based ranking. 
\begin{table}
\caption{Results of entity typing.}
\vspace{-2mm}
\setlength\tabcolsep{2.5pt}
{
\small
\begin{tabular}{c|ccc|ccc}
\newtoprule
\newtoprule
{Datasets} &
\multicolumn{3}{c|}{\YAGO} &
\multicolumn{3}{c}{\DBpedia}
\\
\newmidrule
{Metrics} & 
MRR & Acc. & Hit@3 &
MRR & Acc. & Hit@3
\\
\newtoprule
\newtoprule
{TransE} 	& 0.144 & 7.32 	& 35.26	& 0.503 & 43.67 & 60.78\\ 
{MTransE} 	& 0.689 & 60.87 & 77.64	& 0.672	& 59.87 & 81.32	\\
\newmidrule
\rowcolor{shadecolor} {\texttt{\modelname}-TransE-CG} 	&	{0.829}	& {72.63} 	& 93.35	& {0.828} 	& {70.58} 	& 95.11 \\
\rowcolor{shadecolor} {\texttt{\modelname}-TransE-CT} 	&	{0.843}	& {75.31} 	& {93.18}	& {0.846} 	& {74.41} & {{94.53}} \\
\rowcolor{shadecolor} {\texttt{\modelname}-HATransE-CT}  	& \underline{\textbf{0.897}}	& \underline{\textbf{85.60}} 	& \underline{\textbf{95.91}} 	& \underline{\textbf{0.857}	}	& \underline{\textbf{75.55}}	& \underline{\textbf{95.91}} \\ 
\newmidrule
{DistMult} 	& 0.411 & 36.07 & 55.32	& 0.551 & 49.83 & 68.01\\
\rowcolor{shadecolor} {\texttt{\modelname}-Mult-CG} 	&	0.762	& 62.62	& 87.82 & 0.764	& 60.83	& 91.80 \\
\rowcolor{shadecolor} {\texttt{\modelname}-Mult-CT} 	&	0.805 	& 70.83	& 89.25	& \textbf{0.791}	& 65.30	& \textbf{93.47} \\  
\rowcolor{shadecolor} {\texttt{\modelname}-HAMult-CT}  	& \textbf{0.865} 	& \textbf{81.63} & \textbf{91.83}  & 0.778 & \textbf{69.38} &	85.71	\\

\newmidrule
{HolE} 		& 0.395 & 34.83 & 54.79 & 0.504 & 44.75 & 65.38\\ 
\rowcolor{shadecolor} {\texttt{\modelname}-HolE-CG} 		&	0.777	& 65.30	& 87.89	& 0.784	& 66.75	& 89.37 \\
\rowcolor{shadecolor} {\texttt{\modelname}-HolE-CT} 		&	0.813	& 72.27	& 88.71	& 0.805	& 68.84 & \textbf{91.22} \\
\rowcolor{shadecolor} {\texttt{\modelname}-HAHolE-CT}  	& \textbf{0.888}	& \textbf{83.67}		& {\textbf{93.87}} & \textbf{0.808} & \textbf{72.51} & 89.79\\
\newbottomrule
\newbottomrule 
\end{tabular}
}
\label{tab:type}
\end{table}

\stitle{Results}
Results are reported in Table \ref{tab:type}. All \texttt{\modelname} variants perform significantly better than the baselines. The best \texttt{\modelname} model, i.e. \texttt{\modelname}-TransE-CT, outperforms the best baseline model MTransE by 15.4\% in terms of accuracy and 14.4\% in terms of $MRR$ on \YAGO. 
The improvement on accuracy and $MRR$ are 14.3\% and 14.5\% on \DBpedia\ compared to MTransE.
The results by other baselines confirm that the cross-view links, which apply to all entities and concepts, cannot be properly captured as a regular relation and requires a dedicated representation technique.

Considering different \texttt{\modelname} variants, our observation is that using translation based intra-view model and CT as the cross-view association model (\texttt{\modelname}-TransE-CT) is consistently better than other settings on both datasets. 
It has an average of 4.1\%  performance gain in $MRR$ over \texttt{\modelname}-HolE-CT and \texttt{\modelname}-DistMult-CT, and an average of 2.17\% performance gain in accuracy over the best of the rest variants (\texttt{\modelname}-TransE-CG).
We believe that, compared with similarity-based intra-view models, translation based intra-view model better differentiates between different entities and different concepts in KGs with directed relations and meta-relations in the KB~\cite{pujara2017sparsity}. 
The results by CT-based model variants are generally better than those by CG-based ones.
We believe this is due to two reasons: (i) CT allows the two embedding spaces have different dimensionalties, and hence better characterizes the ontology-view that is smaller and sparser than the instance view;
(ii) As the topological structures of the two views may exhibit some inconsistency, CT adapts well and is less sensitive to such inconsistency than CG.

In terms of different intra-view models, it is also observed that HA intra-view model with CT settings can 
drastically enhance entity typing task and achieve the best performance especially for \YAGO\ with relatively rich ontology, which improves an average of 6.0\% on $MRR$ and  10.5\% in accuracy compared with the default intra-view settings. The reason that the HA technique does not have similar effects on \DBpedia\ is because \DBpedia\ contains a small ontology with much smaller hierarchical structures\footnote{\DBpedia\ contains 164 ontology-view triples for meta-relations with the hierarchical property, while \YAGO\ contains 1,411.}. Comparing the two datasets, our experiments show that, \texttt{\modelname} generally achieves similar accuracies and $MRR$ scores on \YAGO\ and \DBpedia, but slightly better $Hit@3$ on \DBpedia\ due to its smaller candidate space.

Our method opens up a new direction that the learned embedding may help guide labeling entities with unknown types. In Section \ref{subsec:case} and Appendix \ref{sec:sup-discuss}, we provide more experiments and insights on the benefits of representation learning with \texttt{\modelname}.


\subsection{Case Study}\label{subsec:case}
In this section, we provide two case studies for ontology population and entity typing 
for long-tail entities. 

\stitle{Ontology Population} 
By embedding the meta-relations and concepts in the ontology view, the triple completion process can already populate the ontology view with seen meta-relations, by answering the query like (``\textit{Concert}'',``\textit{Related to}'',$?t$) in the KG completion task. Given the top answers of the query, we can reconstruct triples like (``\textit{Concert}'',``\textit{Related to}'',``\textit{Ballet}'') and (``\textit{Concert}'',``\textit{Related} to'',``\textit{Musical}'') with high confidence. 
However, this process does not resolve the zero-shot cases where some concepts may satisfy some meta-relations that have not pre-existed in the vocabulary of meta-relations.
We cannot predict the potentially new meta-relation "is Politician of" directly with triple completion by answering the following query: (``\textit{Office Holder}'', $?r$, 	``\textit{Country}'').\par

\begin{figure}[!ht]
	\centering
	\includegraphics[width=0.85\columnwidth]{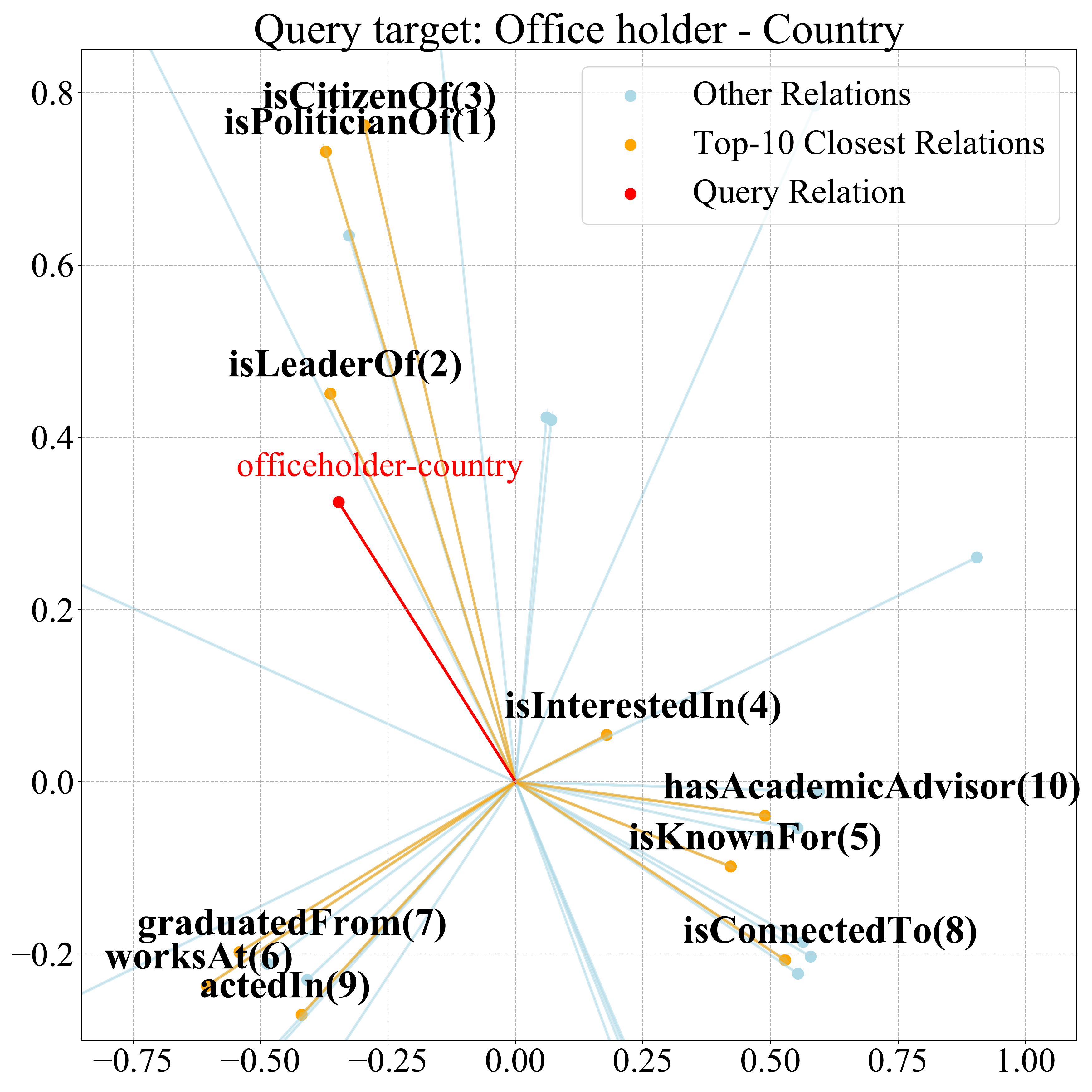}
	\caption{Examples of ontology population by finding the closest relations in the instance view for the query "Office Holder-Country". Top 10 predicted relations are plotted with their ranks.}
	\label{fig:ontopop}
\end{figure}

Our proposed \texttt{\modelname} provides a feasible solution by leveraging the cross-view association model that bridges the two views of the KG, and migrate proper instance-view relations to ontology-view meta-relations.
This is realized by transforming the 
concept embeddings in the query 
to the entity embedding space, and selecting candidate relations from the instance-view.
Considering the previous query (``\textit{Office Holder}'', $?r$, 	``\textit{Country}''), we first find the concept embeddings of ``\textit{Office Holder}'' and ``\textit{Country}'' (denoted as $ \mathbf{c}_{\text{office}} $ and $\mathbf{c}_{\text{country}} $ respectively ), and then transform them to the entity space.
Specifically, for \texttt{\modelname} variants with translational intra-view model, we find the instance-view relations that are closest to $f_{\text{CT}}^{\text{inv}}(\mathbf{c}_{\text{country}}) - f_{\text{CT}}^{\text{inv}}(\mathbf{c}_{\text{office}})$. 
Figure \ref{fig:ontopop} shows the PCA projections of the top 10 relation prediction results for this query. 
The top 3 relations are ``\textit{is Politician of}'', ``\textit{is Leader of}'' and ``\textit{is Citizen of}'', which are all reasonable answers. 
\vspace{-4pt}
\vspace{-8pt}
\begin{table}[htbp]
\centering
\caption{Examples of ontology population from \texttt{\modelname}-TransE-CT. Top 5 Populated Triples with smallest L2-norm distances are provided with reasonable answers bold-faced.}
\vspace{-2mm}
{\small
\begin{tabular}{c|c}
\newtoprule
\newtoprule
 Query & Top 5 Populated Triples with distances\\
\newtoprule
\newtoprule
\multirow{5}{*}{ \tabincell{c}{(scientist,$?r$,\\ university)}} & scientist, \textit{\textbf{graduated from}}, university (0.499)\\
& scientist, \textit{\textbf{isLeaderOf}}, university (1.082)\\
& scientist, \textit{isKnownFor}, university (1.098)\\
& scientist, \textit{created}, university (1.119)\\
& scientist, \textit{\textbf{livesIn}}, university (1.141)\\
\newmidrule
\multirow{5}{*}{ \tabincell{c}{(boxer, $?r$,\\ club)}}  & boxer, \textit{\textbf{playsFor}}, club (1.467) \\
& boxer, \textit{\textbf{isAffiliatedTo}}, club (1.474)  \\
& boxer, \textit{\textbf{worksAt}}, club (1.479)  \\
& boxer, \textit{{graduatedFrom}}, club (1.497)  \\
& boxer, \textit{{isConnectedTo}}, club (1.552)  \\
\newmidrule
\multirow{5}{*}{ \tabincell{c}{(TV station, $?r$, \\country)}}  & TV station, \textit{\textbf{headquarter}}, country (1.221)\\
& TV station, \textit{{parentOrganisation}}, country (1.246) \\
& TV station, \textit{{appointer}}, country (1.253)\\
& TV station, \textit{\textbf{broadcastArea}}, country (1.266) \\
& TV station, \textit{\textbf{principalArea}}, country (1.271) \\
\newmidrule
\multirow{5}{*}{ \tabincell{c}{(scientist, $?r$,\\ scientist)}}  & scientist, \textit{{deputy}}, scientist (0.204)\\
& scientist,\textit{\textbf{doctoralAdvisor}}, scientist (0.218) \\
& scientist, \textit{\textbf{doctoralStudent}}, scientist (0.221)\\
& scientist, \textit{\textbf{relative}}, scientist (0.228)\\
& scientist, \textit{\textbf{spouse}}, scientist (0.230)\\
\newbottomrule
\newbottomrule
\end{tabular}
}
\label{tab:ontopop}
\end{table}

Table \ref{tab:ontopop} 
shows some examples of newly discovered meta-relation facts that have not pre-existed in the ontology views of the two datasets. Five predictions with the highest plausibility (smallest distance) are provided for each query from the ontology-view graph. From these top predictions, we observe that most populated ontology triples migrated from the instance view are meaningful.

\stitle{Long-tail entity typing} In KGs, the frequency of entities and relations often follow a long-tail distribution (Zipf's law). As shown in Figure \ref{fig:yago_ins} and Figure \ref{fig:db_ins} (in Appendix \ref{sec:sup-dataset}), both \YAGO\ and \DBpedia\ discover such a property. Over 75\% of total entities has less than 15 occurrences. 
Those long-tails entities, types and relations are difficult for representation learning algorithms to capture due to being few-shot in training cases.

In this case study, we select the entities with considerably low frequency\footnote{In this experiment, we select entities in \YAGO\ which occurs less than 8 times and entities in \DBpedia\ which occurs less than 3 times.}, which involve around 15\%-30\% of total entities in the instance view of the two KB datasets.
Then, we evaluate the entity typing task for these long-tail entities. 
Table~\ref{tab:long-type} shows the results by the best baselines (DistMult, MTransE) and a groups of our best \texttt{\modelname} variants. 
Similar to our previous observation, \texttt{\modelname} significantly outperforms other baselines. 
Compared with the results in Section \ref{subsec:ex_type}, we observe the depletion of performance for all models, while \texttt{\modelname} variants only have an average of 12.5\% decrease in $MRR$ with CG models and 12.3\% decrease in $MRR$ with CT models while other baselines suffer over 20\% on long-tail entity prediction. There is also an interesting observation that, for long-tails entities, smaller embeddings for both CG ($d_1=d_2=100$) and CT ($d_1=100, d_2=50$) models are beneficial for associated concept prediction. We hypothesize  that this is caused by overfitting on long-tail entities if high dimensionality is used for training without enough training data.

\begin{table}[tbp]
\caption{Results of long-tail entities typing.} 
\vspace{-2mm}
\setlength\tabcolsep{2.5pt}
{\small
\begin{tabular}{c|ccc|ccc}
\newtoprule
\newtoprule
{Datasets} &
\multicolumn{3}{c|}{\YAGO} &
\multicolumn{3}{c}{\DBpedia}
\\
\newmidrule
{Metrics} & 
MRR & Acc. & Hit@3 &
MRR & Acc. & Hit@3
\\
\newtoprule
\newtoprule
{DistMult} 	& 0.156 & 10.89 & 25.33 & 0.219 & 16.48  &  33.71  \\
{MTransE}   & 0.526 & 46.45 & 67.25 & 0.505 & 46.67 & 64.36  \\ 
\newmidrule
\rowcolor{shadecolor}{\texttt{\modelname}-TransE-CG} 	& 0.708 	& 59.97  & 79.80 & 0.741 & 64.45 & 83.05 \\
\rowcolor{shadecolor}{\texttt{\modelname}-TransE-CT} 	& 0.737 & 62.05 & 82.60 & 0.758 & 66.35 & 83.80 \\ 
\rowcolor{shadecolor} {\texttt{\modelname}-HATransE-CT}  	& \textbf{0.802}	& \textbf{69.66} 	& \textbf{87.75} 	& \textbf{0.760}		& \textbf{67.34}	& \textbf{89.79} \\
\newbottomrule
\newbottomrule
\end{tabular}
}
\label{tab:long-type}
\end{table}

In Table \ref{tab:example_longtail}, we include some examples of top 3 predicted categories of long-tail entities by DistMult, MTransE and \texttt{\modelname} (using \texttt{\modelname}-HATransE-CT variant) from \DBpedia, when the instance-view graph and ontology-view graph are relatively sparser. \texttt{\modelname} is still able to make correct predictions of low-frequency entities while other baselines models can only output  inaccurate predictions.

\begin{table}
\centering
\caption{Examples of long-tail entity typing. Top 3 predictions are provided with the correct type bold-faced. }
\vspace{-2mm}
{
\small
\begin{tabular}{c|c|c}
\newtoprule
\newtoprule
Entity & Model & Top 3  Concept Prediction\\
\newbottomrule
\multirow{3}{*} { \tabincell{c}{ Laurence \\ Fishburne }} & DistMult & football team, club, team\\
& MTransE & writer, \textbf{person}, artist\\
& \texttt{\modelname} & \textbf{person}, artist, philosopher\\
\newmidrule
\multirow{3}{*} { \tabincell{c}{ Warangal \\ City }} & DistMult & country, village,\textbf{city}\\
& MTransE & administrative region, \textbf{city}, settlement\\
& \texttt{\modelname} & \textbf{city}, town, country\\
\newmidrule
\multirow{3}{*} { \tabincell{c}{ Royal Victor\\-ian Order }} & DistMult & person, writer, administrative region \\
& MTransE & election, award, \textbf{order}\\
& \texttt{\modelname} & award, \textbf{order}, election\\
\newbottomrule
\newbottomrule
\end{tabular}
}
\label{tab:example_longtail}
\end{table}

\section{Conclusion and Future Work}\label{sec:conclusion}

In this paper, we propose a novel model \texttt{\modelname} aiming to jointly embed real-world entities and ontological concepts. 
We characterize a two-view knowledge base. In the embedding space, our approach jointly captures both structured knowledge of each view, and cross-view links 
that bridges the two views. 
Extensive experiments on the tasks of KG completion and entity typing show that our model \texttt{\modelname} can successfully capture latent features from  both views in KBs, and outperforms various state-of-the-art baselines.

We also point out future directions and improvements. Particularly, instead of optimizing structure loss with triples (first-order neighborhood) locally, we plan to adopt more complex embedding models which leverage information from higher order neighborhood, logic paths or even global knowledge graph structures. We also plan to explore  the alignment on relations and meta-relations like entity-concept.

\vspace{-2pt}
\begin{acks}
The authors would like to thank the anonymous reviewers for their insightful and constructive comments.
This work was partially supported by NIH R01GM115833, U01HG008488, NSF DBI-1565137, DGE-1829071, NSF III-1705169, NSF Career Award 1741634 and Amazon Research Award. 
\end{acks}
\vspace{-2pt}
\bibliographystyle{ACM-Reference-Format}
\bibliography{ref.bib}

\newpage\clearpage
\newpage
\appendix
\section{Ablation Study} \label{sec:sup-discuss}


In this section, we provide some insights on several critical factors that affect the performance of the model. 
These include the embedding dimensionality, sufficiency of cross-view links in training, and the effect of adopting negative sampling in cross-view association models. 

\subsection{Dimensionality}
\begin{figure}
\centering
\begin{subfigure}[b]{1\columnwidth}
   \centering
   \includegraphics[width=1\columnwidth]{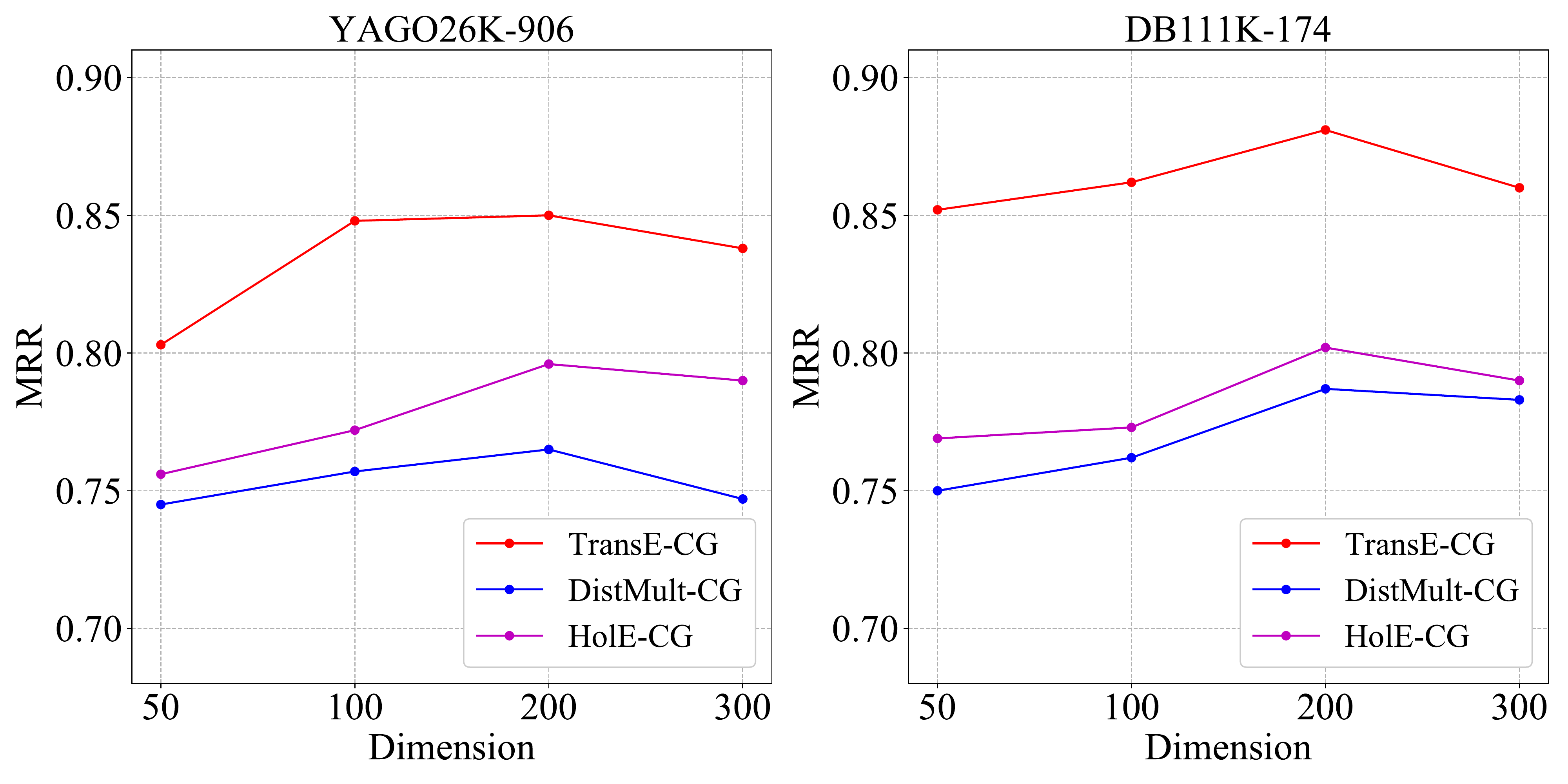}
   \caption{Different dimensions with CG variants}
   \label{fig:dim_CG} 
\end{subfigure}
\begin{subfigure}[b]{1\columnwidth}
   \centering
   \includegraphics[width=1\columnwidth]{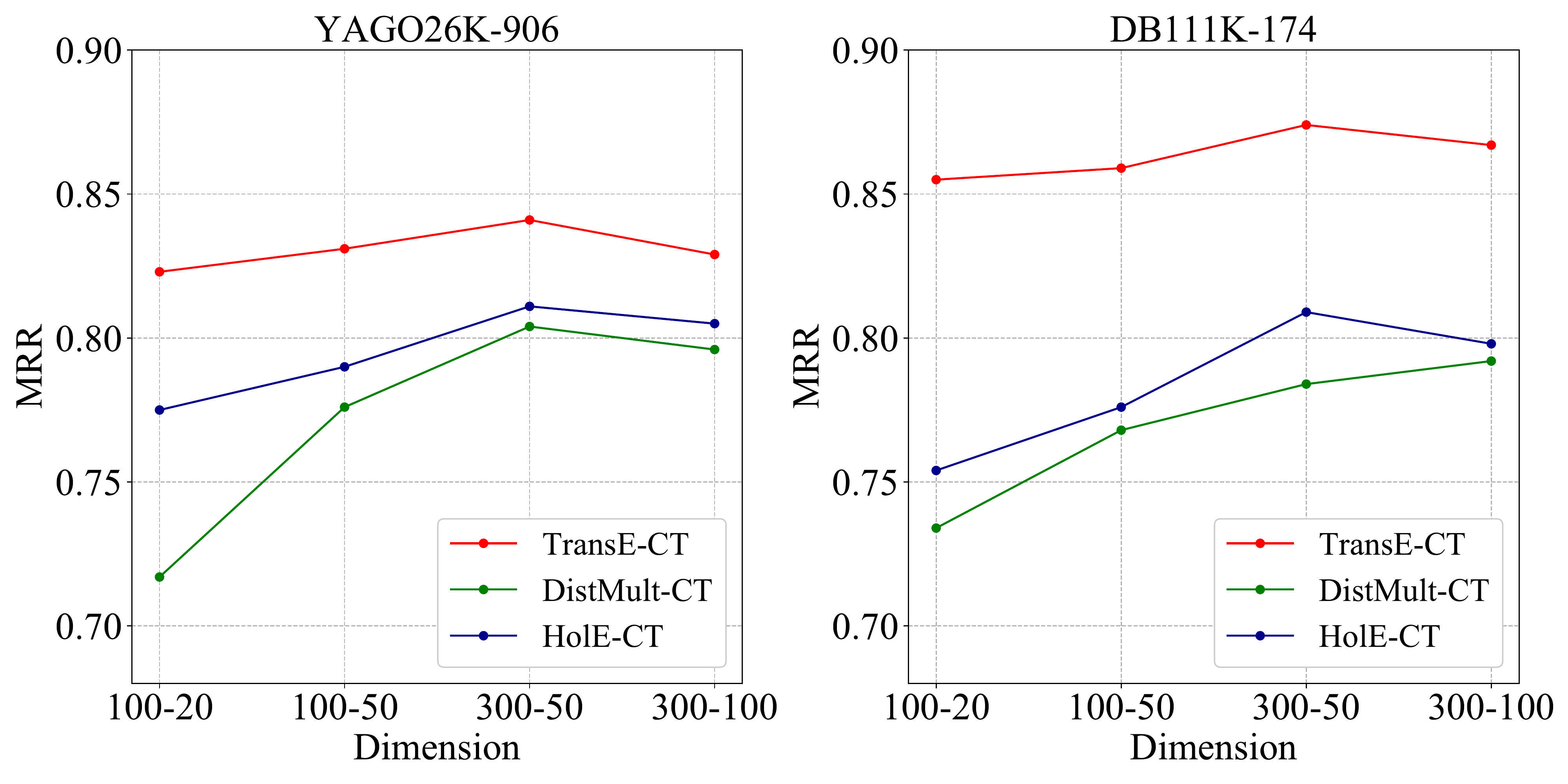}
   \caption{Different dimensions with CT variants}
   \label{fig:dim_CT}
\end{subfigure}
\caption{Performances of entity typing task on both datasets with different entity and concept embedding dimensionalities}
\end{figure}

Dimensionality is a key hyperparameter that affects the quality of the obtained embeddings.
Figure~\ref{fig:dim_CG} shows the $MRR$ of model variants with the CG-based cross-view association according to different embedding dimensions $d$.
It is observed in Figure \ref{fig:dim_CG} that the performance of CG variants are generally improving from $d=50$ to $d=200$, however, after reaching the optimal $d_{\text{opt}}=200$, $MRR$ begins to drop at $d=300$.
Similarly we plot $MRR$ scores for both dataset with CT model variants in Figure \ref{fig:dim_CT}. 

We compare four different dimensionality settings of $(d_1, d_2)$: $(100,20)$,$(100,50)$,$(300,50)$ and $(300,100)$\footnote{$(d_1,d_2)=(100,20)$ denotes that entities are embedded with $d_1=100$ dimensional vectors and concepts are embedded with $d_2=20$ dimensional vectors}.
Most of the \texttt{\modelname} variants achieve their best performance under  the embedding setting $(d_1,d_2)=(300,50)$ rather than $(d_1,d_2)=(300,100)$ (except \texttt{\modelname}-Mult-CT on \DBpedia).
The reason is that, \texttt{\modelname} set with 
low dimensionalities easily falls short of capturing latent features of entities and concepts, while too high dimensionalities lead to overfitting on the ontology view of KG, as well as inefficient training and prediction processes. 

\subsection{Sufficiency of Type Information}
Cross-view links between the instance-view graph and the ontology-view graph are key components, which bridge and enable the information flow between two views to generate embeddings. We also investigate the influence of  cross-view links and their sufficiency in training. 

\begin{figure}[htbp]
    \centering
    \begin{subfigure}[t]{0.5\columnwidth}
        \centering
        \includegraphics[width=1\columnwidth]{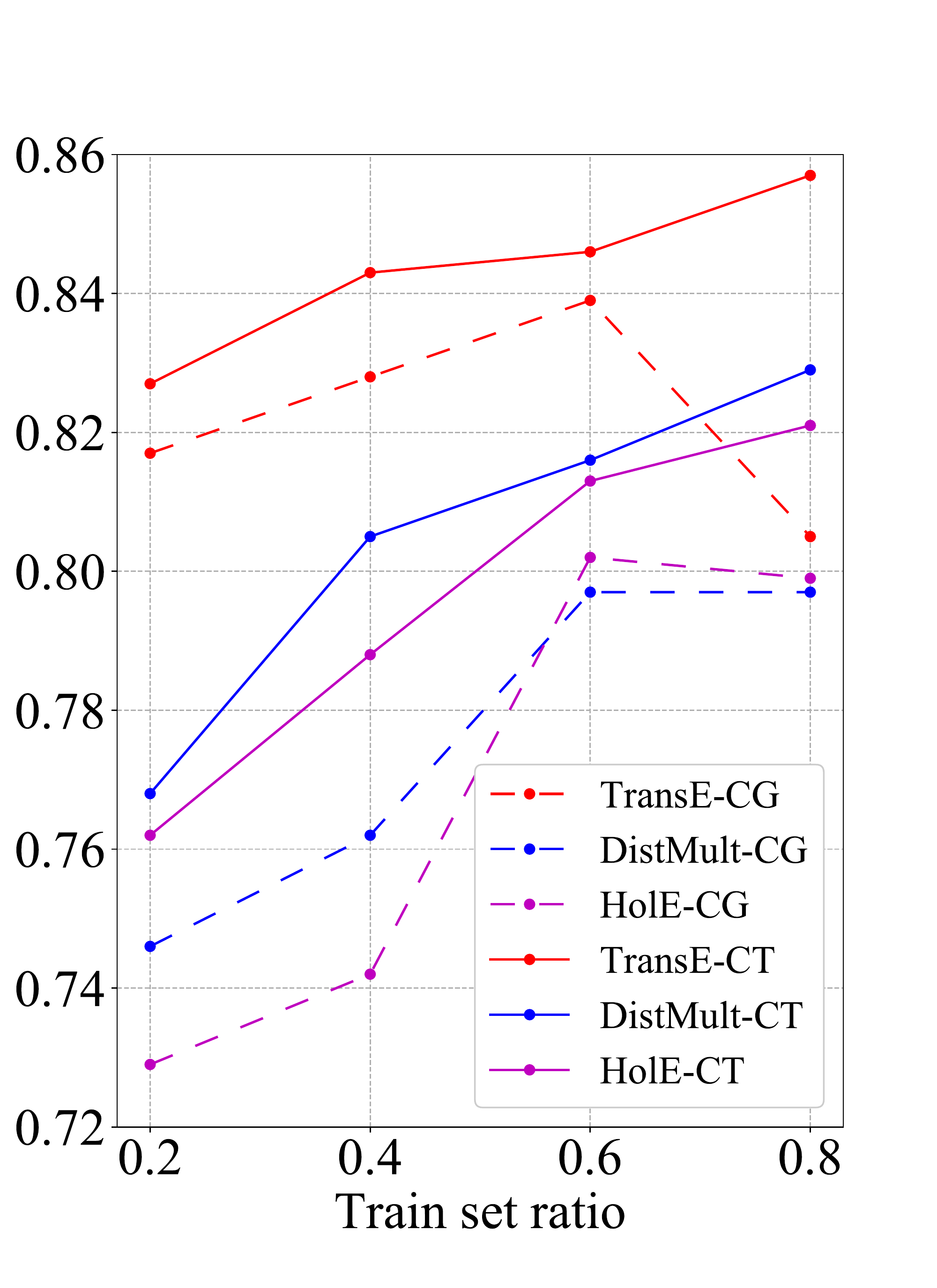}
        \caption{\YAGO}
        \label{fig:ratio_YAGO} 
    \end{subfigure}%
    ~
    \begin{subfigure}[t]{0.5\columnwidth}
        \centering
        \includegraphics[width=1\columnwidth]{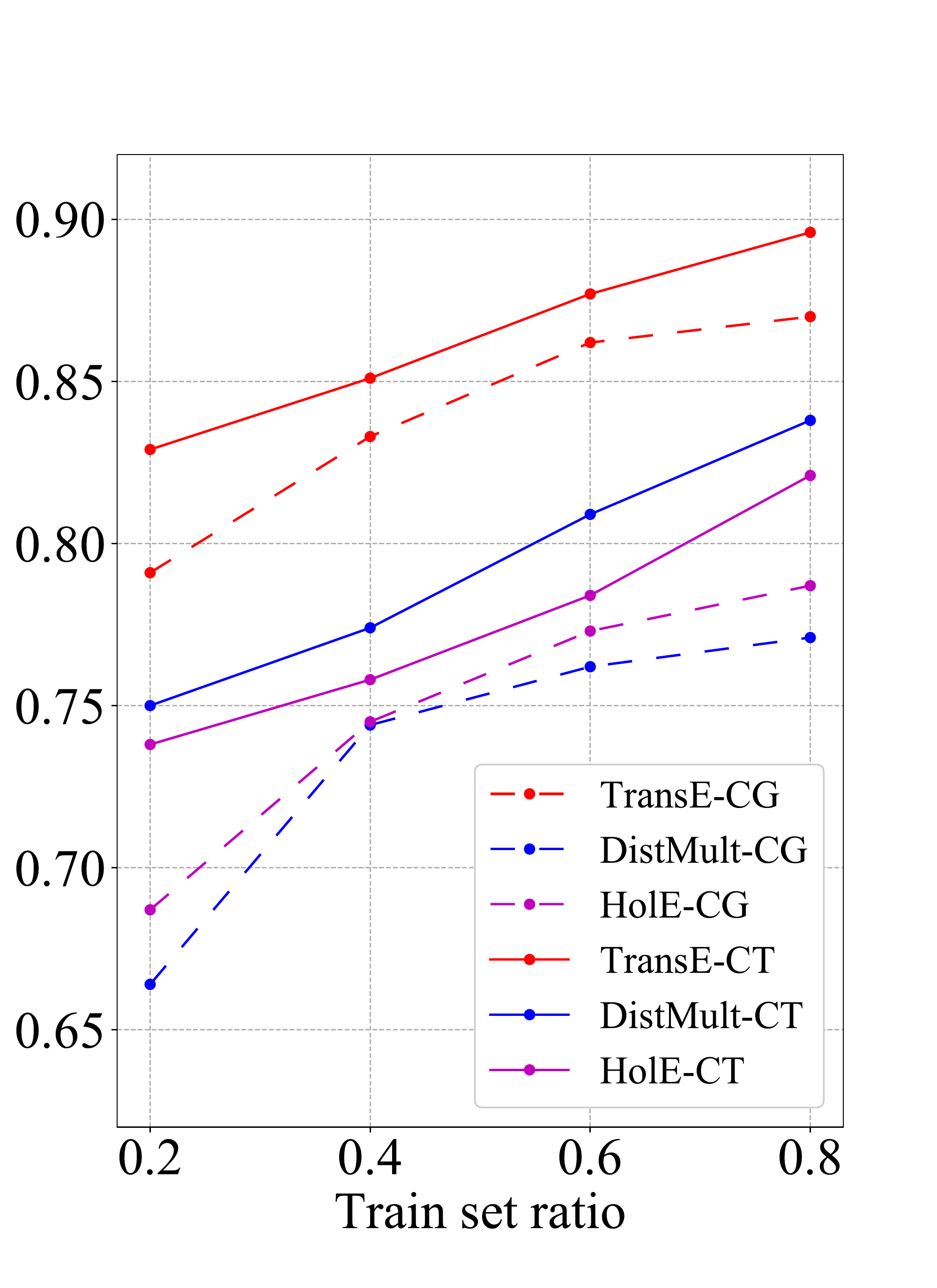}
        \caption{\DBpedia}
        \label{fig:ratio_DB}
    \end{subfigure}
    \caption{The effect of training the model using different proportions of cross-view links on (a) \YAGO\ and (b) \DBpedia}
\label{fig:train-ratio}
\end{figure}

We define the train set ratio $\nu = \{0.2,0.4,0.6,0.8\}$, which means the proportions of the cross-view links that are used for training \texttt{\modelname}.
$MRR$ score is reported in Figure \ref{fig:ratio_YAGO}\ on \YAGO\ and Figure \ref{fig:ratio_DB} on \DBpedia. 
As expected, when the proportion of cross-view links used for training increasing from 20\% to 80\%, the performance improves by 3.2\%  on \YAGO\ and by 2.9\% on \DBpedia\  in terms of $MRR$. 
It is noteworthy that \texttt{\modelname} trained with 20\% cross-view links still outperforms MTransE trained with 60\% cross-view links, which indicates that one advantage of \texttt{\modelname} is its outstanding generalization ability to other untyped entities, given limited knowledge on entity-concept pairs. 

One interesting observation is that, when $\nu$ increases from 0.6 to 0.8, the performance of CG variants does not necessarily improve, while the performance of CT variants still has significant improvements. 
We hypothesize that this is because the strong clustering-based constraint in CG 
can be sensitive to even minor inconsistencies between the topological structures of the two KG views, giving too much supervision. 
CT, on the contrary, 
is more robust against the inconsistency between the two views.
There is a trade-off between the robustness of CT and the efficiency of CG.

\subsection{Effects of Negative Sampling}
Negative sampling is widely applied in the encoding process of a single KG structure~\cite{bordes2013translating,yang2014embedding}.
One interesting question is whether to use negative sampling 
for capturing the cross-view links between two structures,
i.e. to provide corrupted entity-concept pairs such as (``Barack Obama'',``state''). We compare the results of entity typing task by \texttt{\modelname}\ variants with and without cross-view link negative samples in Table \ref{tab:ns}.
It is our finding that there is a significant performance drop if negative sampling is 
disabled in CT, while negative sampling has less effect on CG. We hypothesize that the difference is attributed to the fact that strong clustering-based constraint of CG is already effective in separating irrelevant concepts. 

\begin{figure}[htbp]
    \centering
    \begin{subfigure}[t]{1\columnwidth}
        \centering
        \includegraphics[width=0.95\columnwidth]{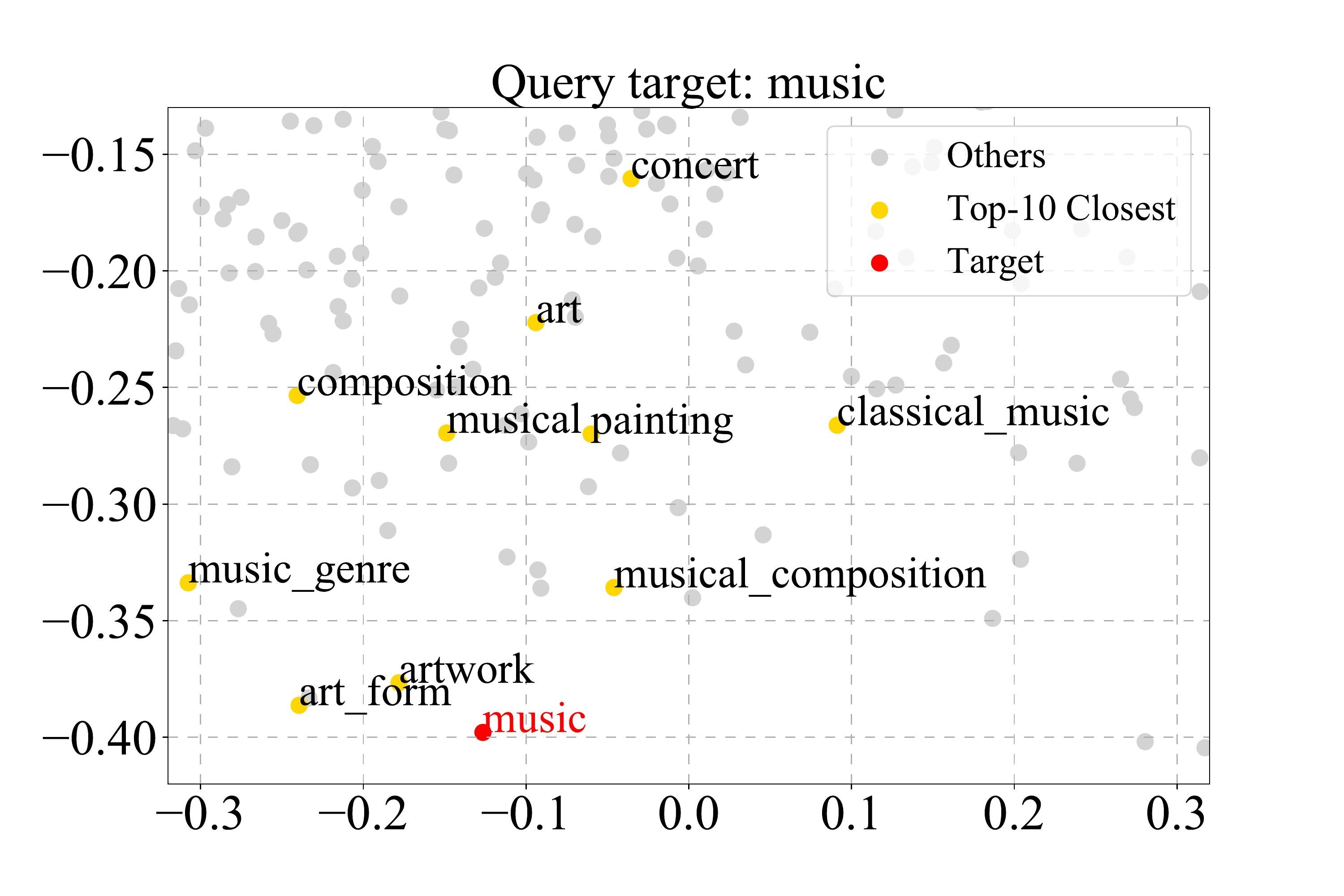}
        \caption{\texttt{\modelname}-TransE-CT (With negative sampling)}
    \end{subfigure}
    \begin{subfigure}[t]{1\columnwidth}
        \centering
        \includegraphics[width=0.95\columnwidth]{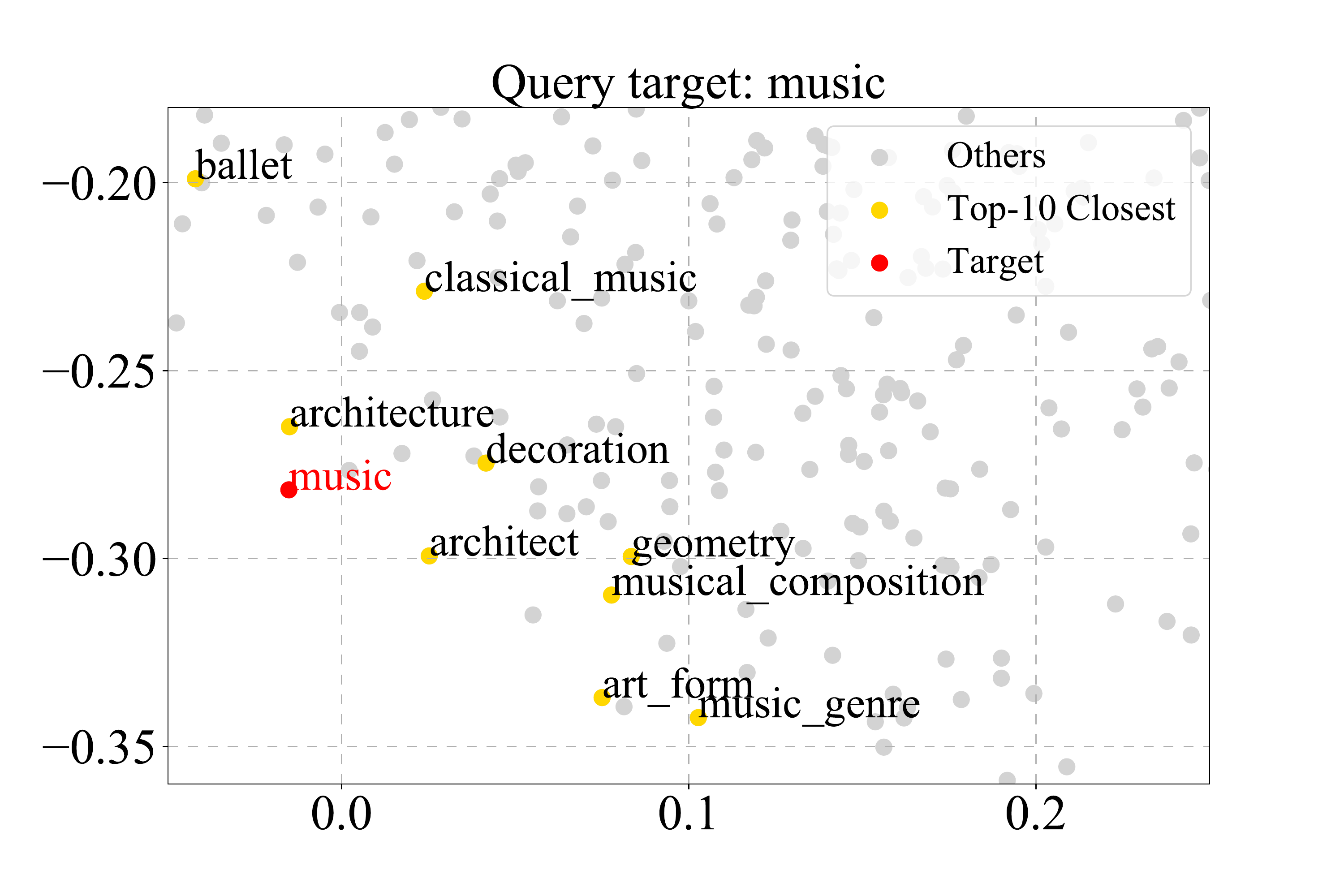}
        \centering
        \caption{\texttt{\modelname}-TransE-CT (Without negative sampling)}
    \end{subfigure}
    \caption{Visualize effects on embeddings of negative sampling on cross-view links}
    \label{fig:effect_ns}
\end{figure}

We show the effects of negative sampling by visualizing the results of one query, which are plotted as PCA projections in Figure \ref{fig:effect_ns}. 
For the displayed query which targets at the concept ``music'', we plot the 10 nearest neighbors of concepts. 
Although related concepts such as ``classic music'', ``concert'' and ``artist movement''  still stay close by ``music'' in both settings, other irrelevant concepts including ``decoration'' and ``architect'' intercept 
in \texttt{\modelname}-TransE-CT without negative sampling. We find such phenomenon frequently exist in the \texttt{\modelname} embeddings trained without negative sampling, which no-doubt impairs the performance of the entity typing task.

\newcolumntype{a}{>{\columncolor{shadecolor}}c}

\begin{table}
\centering
\caption{Effects of negative sampling in type links}\label{tab:ns}
\vspace{-2mm}

{\small
\begin{tabular}{c|ca|ca}
\newtoprule
\newtoprule
{Datasets} &
\multicolumn{2}{c|}{\YAGO} &
\multicolumn{2}{c}{\DBpedia}
\\
\newmidrule
{Setting} & 
W/O NS & W/ NS &
W/O NS & W/ NS
\\
\newtoprule
\newtoprule
{\texttt{\modelname}-TransE-CG} & 0.657	& 0.805	& 0.815		& 0.864 \\
{\texttt{\modelname}-Mult-CG} & 0.627	& 0.762		& 0.761		& 0.797 \\ 
{\texttt{\modelname}-HolE-CG} & 0.682	& 0.777	& 0.783 	& 0.815 \\
\newmidrule
{\texttt{\modelname}-TransE-CT} & 0.501	& 0.847		& 0.667		& 0.883\\
{\texttt{\modelname}-Mult-CT} & 0.490	& 0.829 	& 0.494	& 0.811\\ 
{\texttt{\modelname}-HolE-CT} & 0.508	& 0.821 	& 0.560	& 0.821 \\
\newbottomrule
\newbottomrule
\end{tabular}
}
\end{table}

\section{Datasets} \label{sec:sup-dataset}
We use YAGO26K-906 and DB111K-174, which are extracted from the connected subsets of YAGO~\cite{mahdisoltani2014yago3} and DBpedia~\cite{lehmann2015dbpedia} respectively, for experimental purpose.
The datasets are constructed through the following steps:

\begin{enumerate}
	\item We first filter out all attribute triples, since such triples do not represent the relations of entities or concepts. After randomly sample some relational triples from the rest of the filtered dataset since original YAGO and DBpedia both have large collections of instance-view triples.
	\item After we obtain the entity set of instance view, we extract cross-view alignment of those entities to the ontology view of the two KBs. As a result, a portion of entities are linked to the associated concepts, which are naturally the nodes in the ontology view.
	\item Given all the associated concepts from step (2), we construct the corresponding ontology views base on the intersecting subgraph of the original ontologies.
\end{enumerate}

It is noteworthy that the original YAGO has a taxonomical ontology with only three types of semantic relations, which casts limitation on semantic relations among concepts.
Therefore, we enrich the ontology view of YAGO using the knowledge from ConceptNet~\cite{speer2017conceptnet}, another KB which contains a large collection of meta-relations among concepts.
The concepts in ConceptNet and YAGO are easily aligned by the shared WordNet-based IDs or concept names. 
Consequently, we obtain two datasets that are much larger than FB15K -- the widely adopted instance KG benchmark dataset by many recent works~\cite{bordes2013translating,yang2014embedding,lin2015learning,nickel2016holographic}. \par

\begin{figure}[htbp]
    \centering
    \begin{subfigure}[t]{0.5\columnwidth}
        \centering
        \includegraphics[width=1\columnwidth]{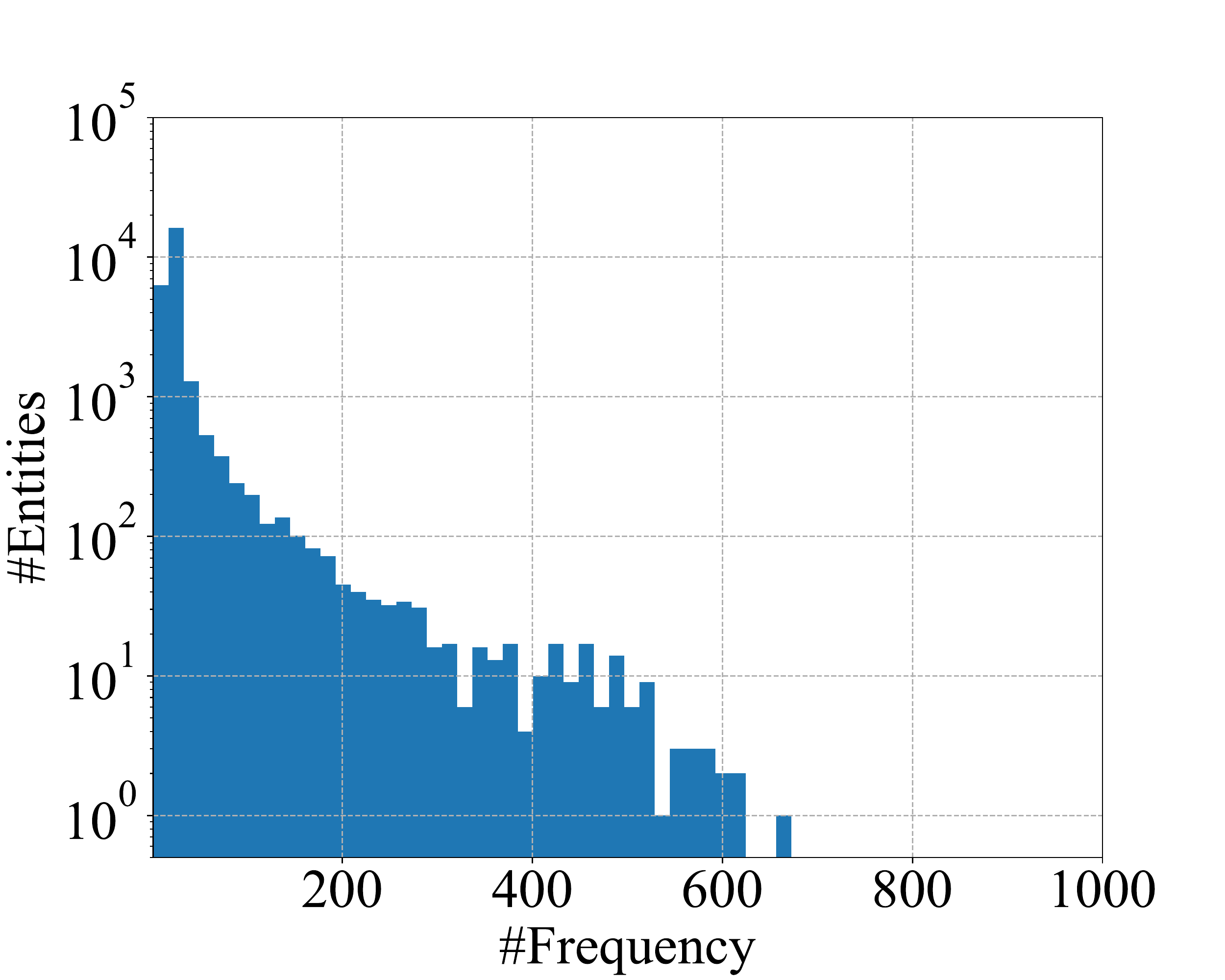}
        \caption{Histogram: \YAGO\ \\ Entity Frequency}
        \label{fig:yago_ins} 
    \end{subfigure}%
    ~
    \begin{subfigure}[t]{0.5\columnwidth}
        \centering
        \includegraphics[width=1\columnwidth]{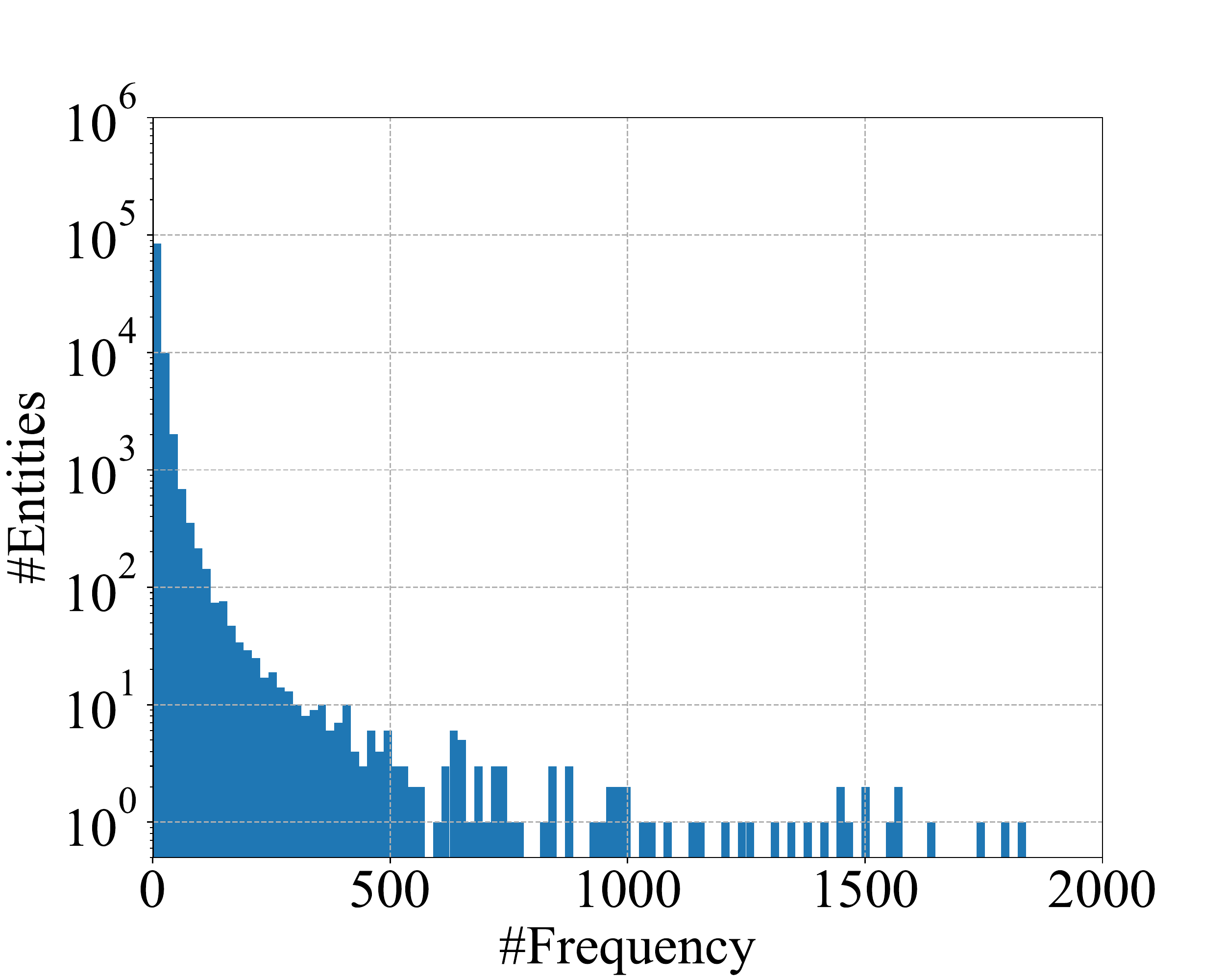}
        \caption{Histogram: \DBpedia\ \\ Entity Frequency}
        \label{fig:db_ins}
    \end{subfigure}
    \caption{Long-tail distribution holds on entity frequency from both \YAGO(a) and \DBpedia(b)}
\label{fig:histogram}
\end{figure}

As stated in Section \ref{subsec:case}, the frequency of entities and relations often follow a long-tail distribution (Zipf's law) in both YAGO26K-906 and DB111K-174 datasets, which is confirmed by the histogram in Figure \ref{fig:histogram}.

\end{document}